\documentclass[acmsmall]{acmart}
\AtBeginDocument{%
  \providecommand\BibTeX{{%
    \normalfont B\kern-0.5em{\scshape i\kern-0.25em b}\kern-0.8em\TeX}}}

\setcopyright{acmlicensed}
\copyrightyear{2018}
\acmYear{2018}
\acmDOI{XXXXXXX.XXXXXXX}

\acmJournal{JACM}
\acmVolume{37}
\acmNumber{4}
\acmArticle{0}
\acmMonth{8}

\usepackage{pifont}
\usepackage{amssymb}
\usepackage{algorithm}
\usepackage{algorithmic}
\usepackage{bibentry}
\usepackage{amsfonts}
\usepackage{subfigure}
\usepackage{multirow}
\usepackage{makecell}
\usepackage{amsmath}
\usepackage{graphicx}
\usepackage{epsfig}
\usepackage{color}
\usepackage{epstopdf}
\usepackage{rotating}
\usepackage{caption}
\usepackage{float}
\usepackage{multicol}
\usepackage{graphicx}
\usepackage{bbding}
\usepackage{algorithm}
\usepackage{algorithmic}
\usepackage{balance}
\usepackage{microtype}
\usepackage{booktabs} 
\usepackage{multirow}
\usepackage{amsmath}
\usepackage{amsmath, bm}

\usepackage{mathtools}
\usepackage{etoolbox}
\usepackage{cases}
\usepackage{enumitem}
\usepackage{xcolor}
\usepackage{xcolor}
\usepackage{color}
\usepackage{amssymb}

\usepackage{cases}

\usepackage{amsmath,amssymb,amsfonts}
\usepackage{algorithmic}
\usepackage{graphicx}
\usepackage{textcomp}

\usepackage{mathtools}

\usepackage{amssymb}
\usepackage{amsthm}
\usepackage{color}
\usepackage{multirow}
\usepackage{amssymb}

\usepackage{cases}
\usepackage{amsmath,amssymb,amsfonts}
\usepackage{algorithmic}
\usepackage{graphicx}
\usepackage{textcomp}
\usepackage{xcolor}
\usepackage{color}

\usepackage{adjustbox}
\usepackage{lipsum}
\usepackage{enumitem}
\usepackage{tikz}
\usepackage{fontawesome}
\usepackage{scalerel}
\usepackage[edges]{forest}
\definecolor{hidden-draw}{RGB}{0,0,0}
\definecolor{hidden-pink}{RGB}{168,191,143}

\usepackage{colortbl}  
\usepackage{array} 
\usepackage{tabularx}
\usepackage{framed}

\newcommand{\red}[1]{\textcolor{red}{#1}}
\definecolor{brown}{RGB}{139,64,0}
\definecolor{pink}{RGB}{255,170,182}
\definecolor{purple}{RGB}{160,32,240}
\definecolor{g}{RGB}{217,244,212}
\definecolor{p}{RGB}{237,203,201}
\definecolor{b}{RGB}{237,244,252}
\definecolor{lightgreen}{RGB}{44,160,44}
\definecolor{orange}{RGB}{253,229,206}
\definecolor{myorange}{RGB}{247,171,145}
\definecolor{green}{RGB}{213,232,212}
\definecolor{slightblue}{RGB}{189,215,238}
\newcommand{\lightgreen}[1]{\textcolor{lightgreen}{#1}}

\usepackage{amssymb}

\usepackage{cases}
 
\usepackage{amsmath,amssymb,amsfonts}
\usepackage{algorithmic}
\usepackage{graphicx}
\usepackage{textcomp}
\usepackage{xcolor}

\begin{document}

\title{Graph Machine Learning in the Era of Large Language Models (LLMs)}

\author{Shijie Wang}
\authornote{Both authors contributed equally to this research.}
\affiliation{%
  \institution{The Hong Kong Polytechnic University}
  \country{Hong Kong}}
\email{shijie.wang@connect.polyu.hk}

\author{Jiani Huang}
\authornotemark[1]
\affiliation{%
  \institution{The Hong Kong Polytechnic University}
  \country{Hong Kong}}
\email{jia-ni.huang@connect.polyu.hk}

\author{Zhikai Chen}
\affiliation{%
  \institution{Michigan State University}
  \country{USA}}
\email{chenzh85@msu.edu}

\author{Yu Song}
\affiliation{%
  \institution{Michigan State University}
  \country{USA}}
\email{songyu5@msu.edu}

\author{Wenzhuo Tang}
\affiliation{%
  \institution{Michigan State University}
  \country{USA}}
\email{tangwen2@msu.edu}

\author{Haitao Mao}
\affiliation{%
  \institution{Michigan State University}
  \country{USA}}
\email{haitaoma@msu.edu}

\author{Wenqi Fan}
\authornote{Corresponding authors}
\affiliation{%
  \institution{The Hong Kong Polytechnic University}
  \country{Hong Kong}}
\email{wenqifan03@gmail.com}

\author{Hui Liu}
\affiliation{%
  \institution{Michigan State University}
  \country{USA}}
\email{liuhui7@msu.edu}

\author{Xiaorui Liu}
\affiliation{%
  \institution{North Carolina State University}
  \country{USA}}
\email{xliu96@ncsu.edu}

\author{Dawei Yin}
\affiliation{%
  \institution{Baidu Inc}
  \country{China}}
\email{yindawei@acm.org}

\author{Qing Li}
\authornotemark[2]
\affiliation{%
  \institution{The Hong Kong Polytechnic University}
  \country{Hong Kong}}
\email{csqli@comp.polyu.edu.hk}

\renewcommand{\shortauthors}{Fan and Wang, et al.}

\begin{abstract}
Graphs play an important role in representing complex relationships in various domains like social networks, knowledge graphs, and molecular discovery. With the advent of deep learning, Graph Neural Networks (GNNs) have emerged as a cornerstone in Graph Machine Learning (Graph ML), facilitating the representation and processing of graphs. Recently, LLMs have demonstrated unprecedented capabilities in language tasks and are widely adopted in a variety of applications such as computer vision and recommender systems. This remarkable success has also attracted interest in applying LLMs to the graph domain. Increasing efforts have been made to explore the potential of LLMs in advancing Graph ML's generalization, transferability, and few-shot learning ability. Meanwhile, graphs, especially knowledge graphs, are rich in reliable factual knowledge, which can be utilized to enhance the reasoning capabilities of LLMs and potentially alleviate their limitations such as hallucinations and the lack of explainability. Given the rapid progress of this research direction, a systematic review summarizing the latest advancements for Graph ML in the era of LLMs is necessary to provide an in-depth understanding to researchers and practitioners. Therefore, in this survey, we first review the recent developments in Graph ML. We then explore how LLMs can be utilized to enhance the quality of graph features, alleviate the reliance on labeled data, and address challenges such as graph Heterophily and out-of-distribution (OOD) generalization. Afterward, we delve into how graphs can enhance LLMs, highlighting their abilities to enhance LLM pre-training and inference. Furthermore, we investigate various applications and discuss the potential future directions in this promising field.

\end{abstract}





\maketitle

\section{Introduction}
\label{Introduction}
Graph data are widespread in many real-world applications~\cite{mao2023revisiting,hashemi2024comprehensive}, including social graphs, knowledge graphs, and recommender systems~\cite{fan2023adversarial,wu2022disentangled,chen2022knowledge}. 
Typically, graphs consist of nodes and edges, e.g., in a social graph, nodes represent users and edges represent relationships~\cite{derr2020epidemic,ma2021deep}. 
In addition to the topological structure, graphs tend to possess various features of nodes, such as textual description, which provide valuable context and semantic information about nodes.
To effectively model the graph, \emph{Graph Machine Learning (Graph ML)} has garnered significant interest.
With the advent of deep learning (DL), Graph Neural Networks (GNNs) have become a critical technique in Graph ML due to their message-passing mechanism.
This mechanism allows each node to obtain its representation by recursively receiving and aggregating messages from neighboring nodes~\cite{wu2020comprehensive,zhang2024linear}, thereby capturing the high-order relationships and dependencies within the graph structure. To mitigate the reliance on supervised data, many research focused on developing self-supervised Graph ML methods to advance GNNs to capture transferable graph patterns, enhancing their generalization capabilities across various tasks~\cite{you2020graph,zeng2021contrastive,xu2022ccgl,qiu2020gcc}. 
Given the exponential growth of applications of graph data, researchers are actively working to develop more powerful Graph ML methods.

Recently, Large Language Models (LLMs) have started a new trend of AI and have shown remarkable capabilities in natural language processing (NLP)~\cite{radford2018improving,devlin2018bert}.
With the evolution of these models, LLMs are not only being applied to language tasks but also showcasing great potential in various applications such as CV~\cite{zhou2020unified}, and Recommender System~\cite{fan2023recommenderb}.
The effectiveness of LLMs in complex tasks is attributed to their extensive scale in both architecture and dataset size. For example, GPT-3~\cite{brown2020language} with 175 billion parameters demonstrates exciting capabilities by generating human-like text, answering complex questions, and coding. 
Furthermore, LLMs are able to grasp extensive general knowledge and sophisticated reasoning due to their vast training datasets.
Therefore, their abilities in linguistic semantics and knowledge reasoning enable them to learn semantic information. 
Additionally, LLMs exhibit emergence abilities, excelling in new tasks and domains with limited or no specific training. This attribute is expected to provide high generalisability across different downstream datasets and tasks even in few-shot or zero-shot situations.
Thus, given their advantages, leveraging the capabilities of LLMs in Graph Machine Learning (Graph ML) has gained increasing interest.

\begin{figure}[tb]
 \centering
\subfigure{\includegraphics[width=0.6\linewidth]{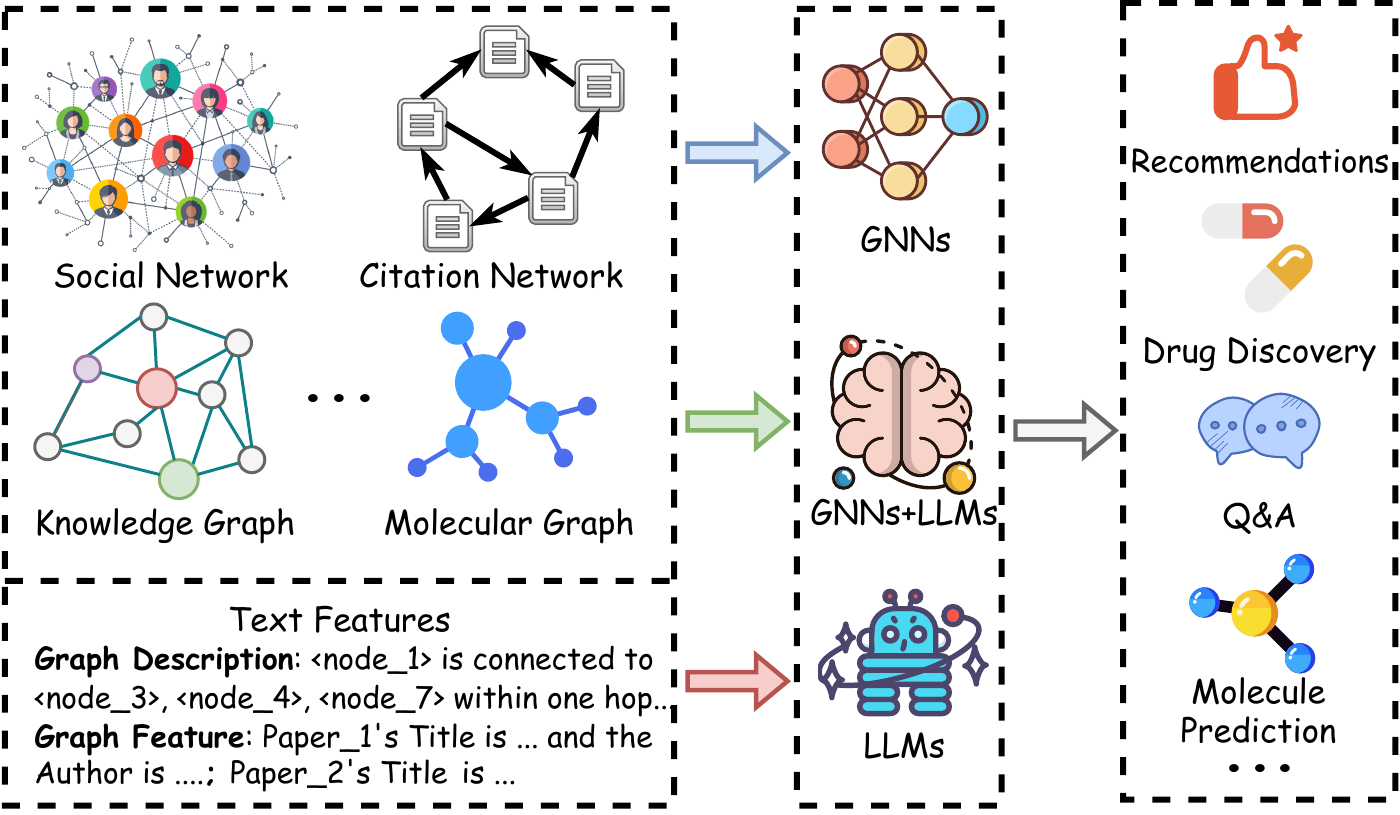}}
\vskip -0.1in
\caption{Illustration of the application of Large Language Models (LLMs) in graph machine learning. The integration of LLMs with Graph Neural Networks (GNNs) is utilized to model an extensive range of graph data across various downstream tasks. }
\label{fig:Illustrate}
\vskip -0.2in
\end{figure}

Currently, many efforts have been made to explore the potential of LLMs in advancing Graph ML. 
By exploiting the ability of LLMs, it is expected to enhance the capability and generalizability of Graph ML on a variety of graph-related tasks.
Figure~\ref{fig:Illustrate} demonstrates an example of integrating LLMs and GNNs for various graph tasks. 
More specifically, some methods leverage LLMs to alleviate the reliance of vanilla Graph ML on labeled data, where they make inferences based on implicit and explicit graph structure information~\cite{ye2023natural,guo2023gpt4graph,tang2023graphgpt}.
For instance, InstructGLM~\cite{ye2023natural} fine-tunes models like LlaMA~\cite{touvron2023llama} and T5~\cite{raffel2020exploring} by serializing graph data as tokens and encoding structural information about the graph to solve graph tasks. 
Meanwhile, to overcome the challenge of feature quality, some methods further employ LLMs to enhance the quality of graph features~\cite{duan2023simteg,chen2023exploring,chien2021node}.
For example, SimTeG~\cite{duan2023simteg} fine-tunes LLMs on textual graphs datasets to obtain textual attribute embeddings, which are then utilized to augment the GNN for various downstream tasks.
Additionally, some studies explore using LLMs to address challenges such as heterogeneity~\cite{zhao2023graphtext} and OOD~\cite{chen2023exploring} of graphs.

On the other hand, although LLM achieves great success in various fields, it still faces several challenges, including hallucinations, actuality awareness, and lacking explainability~\cite{logan2019barack,bubeck2023sparks,zhao2024explainability}. Graphs, especially knowledge graphs, capture extensive high-quality and reliable factual knowledge in a structured format~\cite{chen2022knowledge}. 
Therefore, incorporating graph structure into LLMs could improve the reasoning ability of LLMs and mitigate these limitations.
To this end, efforts have been made to explore the potential of graphs in augmenting LLMs' explainability~\cite{luo2023reasoning,wang2023keqing} and mitigating hallucination~\cite{guan2023mitigating,li2023verifiable}.
Given the rapid evolution and significant potential of this field, a thorough review of recent advancements in graph applications and Graph ML in the era of LLMs is imperative.

\begin{figure*}[tb]
 \centering
\subfigure{\includegraphics[width=0.95\linewidth]{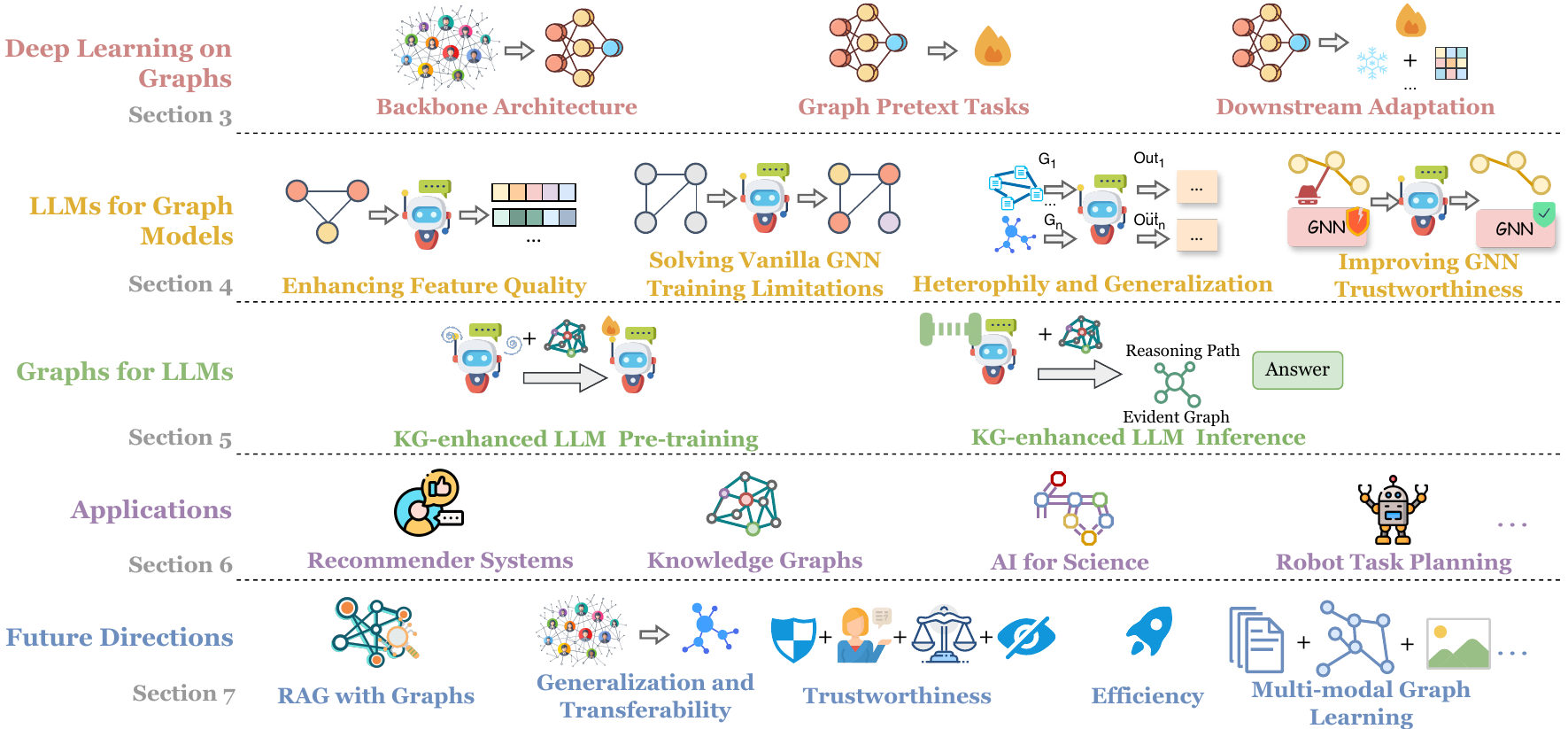}}
\vskip -0.1in
\caption{The outline of our survey. \textbf{Section~\ref{sec:graph foundation model} Deep Learning on Graphs} explores the development of DNN-based methods, focusing on the Backbone Architecture, Graph Pretext Tasks, and Downstream Adaption three aspects. \textbf{Section~\ref{sec:LLM-enhanced Graph} LLMs for Graph
Models} explore how current LLMs help the current Graph ML by Enhancing Feature Quality, Solving Vanilla GNN Training Limitations, Heterophily and Generalization, and Improving GNN Trustworthiness four aspects. 
\textbf{Section~\ref{sec:Graph-enhanced LLMs} Graph for LLMs} focuses on Knowledge Graph(KG)-enhanced LLM Pre-training and KG-enhanced LLM Inference. \textbf{Section~\ref{sec:application} Applications} presents various applications, including Recommender System, Knowledge Graph, AI for Science, and Robot Task Planning. \textbf{Section~\ref{sec:future_work} Future Directions} discusses potential future directions for LLMs in graph machine learning from the Retrieval-Augmented
Generation with Graphs, Generalization and Transferability, Trustworthiness, Efficiency and Multi-modal Graph Learning.}
\label{fig:outline}
\vskip -0.2in
\end{figure*}

\begin{figure*}[tb]
 \centering
\subfigure{\includegraphics[width=0.95\linewidth]{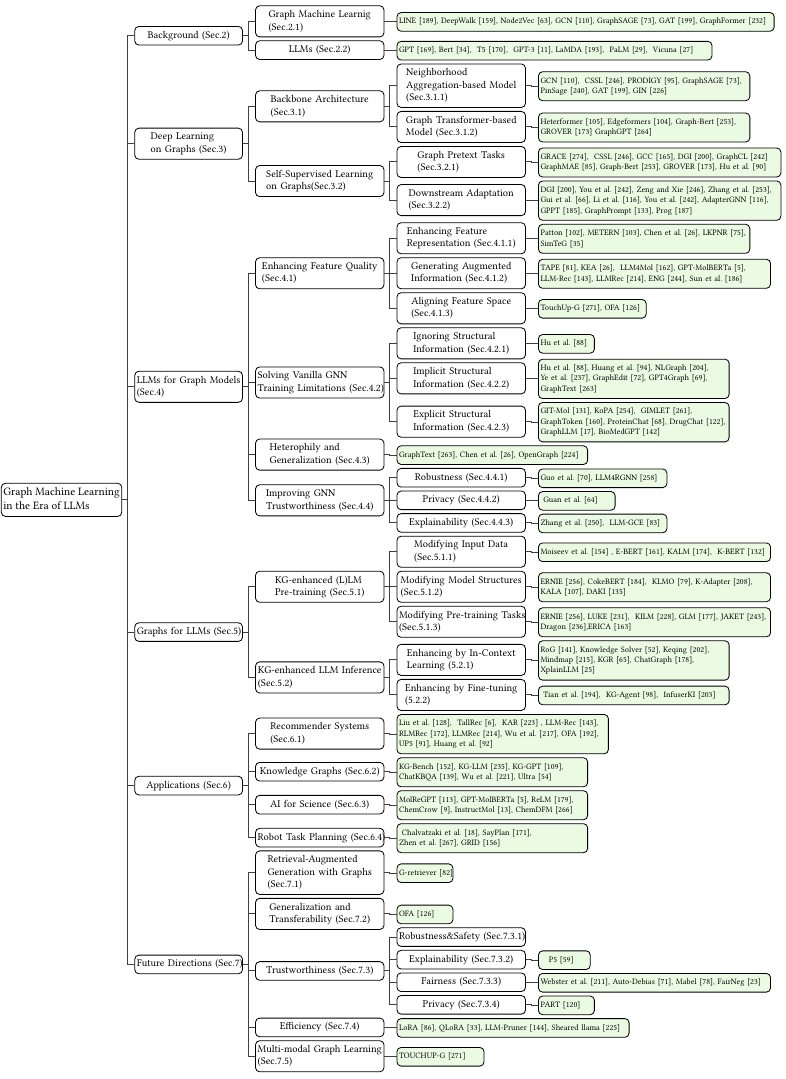}}
\caption{Summary of relevant methods for each section in our survey.}
\label{fig:ref tree}
\vskip -4mm
\end{figure*}

Therefore, in this survey, we aim to provide a comprehensive review of Graph ML in the era of LLMs. 
The outline of the survey is shown in Figure~\ref{fig:outline}: Section~\ref{sec:related work} reviews work related to graph machine learning and LLMs.
Section~\ref{sec:graph foundation model} introduces the deep learning methods on graphs, which focus on various GNN models and self-supervised methods.
Subsequently, the survey delves into how LLMs can be used to enhance Graph ML in Section~\ref{sec:LLM-enhanced Graph} and how graphs can be adopted for augmenting LLMs in Section~\ref{sec:Graph-enhanced LLMs}.
Finally, some applications and potential future directions for Graph ML in the era of LLMs are discussed in Section~\ref{sec:application} and Section~\ref{sec:future_work}, respectively. The relevant literature for each chapter is presented in Figure~\ref{fig:ref tree}. Our main contributions can be summarized as follows:

\begin{itemize}[leftmargin=*]
\item We detail the evolution from early graph learning methods to the latest Graph ML in the era of LLMs;
\item We provide a comprehensive analysis of current LLMs enhanced Graph ML methods, highlighting their advantages and limitations, and offering a systematic categorization;
\item We thoroughly investigate the potential of graph structures to address the limitations of LLMs;
\item We explore the applications and prospective future directions of Graph ML in the era of LLMs, and discuss both research and practical applications in various fields.
\end{itemize}

Concurrent to our survey, Wei \textit{et al.}~\cite{wei2022graph} review the development of graph learning. 
Zhang \textit{et al.}~\cite{zhang2023large} provide a prospective review of large graph models. 
Jin \textit{et al.}~\cite{jin2023large} and Li \textit{et al.}~\cite{li2023survey} review different techniques for pre-training language models (in particular LLMs) on graphs and applications to different types of graphs, respectively.
Liu \textit{et al.}~\cite{liu2023towards} review the Graph Foundation Models according to the pipelines.
Mao \textit{et al.}~\cite{mao2024graph} focus on the fundamental principles and discuss the potential of GFMs.
Different from these concurrent surveys, our survey provides a more comprehensive review with the following differences: (1) we present a more systematic review of the development of Graph Machine Learning and further exploration of LLMs for Graph ML;
(2) we present a more comprehensive and fine-grained taxonomy of recent advancements of Graph ML in the era of LLMs;
(3) we delve into the limitations of recent Graph ML, and provide insights into how to overcome these limitations from LLM's perspective;
(4) we further explore how graphs can be used to augment LLMs; and 
(5) we thoroughly summarize a broad range of applications and present a more forward-looking discussion on the challenges and future directions.

\section{Background}
\label{sec:related work}
In this section, we briefly review some background in the fields of graph machine learning and large language models.

\subsection{Graph Machine Learning}
\label{sec:graph ml}
As one of the most active fields in artificial intelligence, graph learning has attracted considerable attention to its capability to model complex relationships and structures in data represented as graphs~\cite{wang2023fast}. 
Nowadays, it has been widely adopted in various applications, including social network analysis~\cite{wang2024multi}, protein detection~\cite{tsubaki2019compound}, recommender systems~\cite{fan2019graph,fan2020graph}, etc.

The initial phases of graph learning typically use Random Walks, which is a foundational method for exploring graph structures. This technique involves a stochastic process of moving from one node to another within a graph, which is instrumental in understanding node connectivity and influence within networks. Building upon Random Walks, Graph Embedding methods aim to represent nodes (or edges) as low-dimensional vectors while preserving graph topology and node relationships. Representative methods such as LINE~\cite{tang2015line}, DeepWalk~\cite{perozzi2014deepwalk}, and Node2Vec~\cite{grover2016node2vec} leverage Random Walks to learn node representations, capturing local structures and community information effectively.

Due to the exceptional representation learning and modeling capabilities, GNNs bolstered by deep learning have brought significant advances in graph learning. For example, GCNs~\cite{kipf2016semi} introduce convolutional operations to graph data, enabling effective aggregation of neighborhood information for each node, thus enhancing node representation learning.
GraphSAGE~\cite{hamilton2017inductive} learns a function to aggregate information from a node's local neighborhood in an inductive setting, allowing efficient embedding generation for unseen nodes.
GAT~\cite{velivckovic2017graph} further advances GNNs by integrating attention mechanisms, assigning varying weights to nodes in a neighborhood, thereby sharpening the model’s ability to focus on significant nodes. 
Inspired by the success of transformers~\cite{vaswani2017attention} in NLP and CV, several studies~\cite{li2018graph,yun2019graph,baek2021accurate,hu2020heterogeneous,zhang2020graph} adopt self-attention mechanisms to graph data, providing a more global perspective of graph structures and interactions. 
Recent works~\cite{yang2021graphformers,ying2021transformers,mitheran2021introducing,kreuzer2021rethinking,dwivedi2020generalization} further leverage transformer architectures to enhance graph data modeling. For example, GraphFormer~\cite{yang2021graphformers} integrates GNN within each layer in the transformer, enabling simultaneous consideration of textual and graph information.

The advancements in LLMs have given rise to graph learning. Recent works~\cite{guo2023gpt4graph,qian2023can,chen2023exploring,ye2023natural,chen2024graphwiz} apply techniques from these advanced language models like LLaMA~\cite{touvron2023llama} or ChatGPT to graph data, resulting in models capable of understanding and handling graph structures in a manner similar to natural language processing. A typical approach, GraphGPT~\cite{tang2023graphgpt}, tokenizes graph data for insertion into LLMs (i.e. Vicuna~\cite{chiang2023vicuna} and LLaMA~\cite{touvron2023llama}) thus providing a powerful generalization capability. GLEM~\cite{zhao2022learning} further integrates the graph models and LLMs, specifically DeBERTa~\cite{he2020deberta}, within a variational Expectation-Maximization (EM) framework. It alternates between updating LLM and GNN in the E-step and M-step, thereby scaling efficiently and improving effectiveness in downstream tasks.

\subsection{Large Language Models (LLMs)}
\label{sec:llms}
LLMs represent a significant breakthrough in the field of artificial intelligence, ushering in a new era of NLP. 
Characterized by their extensive scale, LLMs are trained on billions of parameters using extensive textual datasets, which equips them to excel in comprehending and generating natural language. The landscape of pre-trained language models is diverse, such as GPT (Generative Pre-trained Transformer)~\cite{radford2018improving}, BERT (Bidirectional Encoder Representations from Trans-
formers)~\cite{devlin2018bert} and T5  (Text-To-Text Transfer Transformer)~\cite{raffel2020exploring}. These models can broadly fall into three categories: encoder-only, decoder-only, and encoder-decoder models. Encoder-only models, such as BERT, specialize in understanding and interpreting language. In contrast, decoder-only models like GPT excel in generating coherent and contextually relevant text. Encoder-decoder models, like T5, combine both abilities, efficiently performing various NLP tasks from translation to summarization.

As an encoder-only model, BERT introduces a paradigm in NLP with its innovative bi-directional attention mechanism, which analyzes text from both directions simultaneously, unlike its predecessors like transformer which processed text in a single direction (either left-to-right or right-to-left).
This feature allows BERT to attain a comprehensive context understanding, significantly improving its language nuance comprehension. 
On the other hand, decoder-only models such as GPT, including variants like ChatGPT, utilize a unidirectional self-attention mechanism. This design makes them particularly effective in predicting subsequent words in a sequence, thus excelling in tasks like text completion, creative writing, and code generation. 
Additionally, as an encoder-decoder model, T5 uniquely transforms a variety of NLP tasks as text generation problems. 
For example, it reframes sentiment analysis from a classification task to a text generation task, where input like "Sentiment: Today is sunny" would prompt T5 to generate an output such as "Positive". This text-to-text approach underscores T5's versatility and adaptability across diverse language tasks.

The evolution of LLMs has seen the emergence of advanced models like GPT-3~\cite{brown2020language}, LaMDA~\cite{thoppilan2022lamda}, PaLM~\cite{chowdhery2022palm}, and Vicuna~\cite{chiang2023vicuna}. These models represent significant advances in NLP, distinguished by their enhanced capabilities in comprehending and generating complex, fine-grained language. 
Their training methods are usually more sophisticated, involving larger datasets and more powerful computational resources. This scaling up has led to unprecedented language understanding and generation capabilities, exhibiting emergent properties such as in-context learning (ICL), adaptability, and flexibility.
Furthermore, recent advancements demonstrate the successful integration of LLMs with other models, like recommender system~\cite{zhao2024recommender}, reinforcement learning (RL)~\cite{shinn2023reflexion}, GNNs~\cite{zhang2023graph,duan2023simteg,mavromatis2023train,jiang2023graphologue}. 
This integration enables LLMs to tackle both traditional and novel challenges, proposing prospective avenues for applications.

Recently, LLMs have found applications in diverse sectors like chemistry~\cite{castro2023large, bran2023chemcrow}, education~\cite{kasneci2023chatgpt,gilson2023does}, and finance~\cite{wu2023bloomberggpt,yang2020finbert}. 
In these fields, they contribute to various tasks from data analysis to personalized learning. 
Particularly, LLMs exhibit great potential in graph tasks such as graph classification and link prediction, demonstrating their versatility and broad applicability.
Specifically, several studies like Simteg~\cite{duan2023simteg}, GraD~\cite{mavromatis2023train}, Graph-Toolformer~\cite{zhang2023graph}, and Graphologue~\cite{jiang2023graphologue} have notably advanced graph learning.
These models utilize LLMs for textual graph learning, graph-aware distillation, and graph reasoning, illustrating the potential of LLMs in enhancing the understanding of and interaction with complex graph structures.

\begin{table*}[t]
\centering
\caption{A comparison of various DNN-based models. We present Models and their Architecture, Pretext Task, Adaptation Method, and Downstream Tasks. 
\textbf{URL} in \textbf{Adaptation Method} indicates Unsupervised Representation Learning.}
\label{tab:Graph_foundation_model}
\resizebox{1\columnwidth}{!}{
\begin{tabular}{ccccc}
\toprule
 \multirow{1}{*}{Adaptation Method} & {Model} & {Architecture} &  {Pretext Task} &  {Downstream Tasks} \\
    \midrule
URL & DGI~\cite{velivckovic2018deep} &
GNN & Contrastive Learning &  Node \\
URL & GRACE~\cite{zhu2020deep} & GNN & Contrastive Learning & Node \\

URL & GraphMAE~\cite{hou2022graphmae}& GNN & Graph Generation &Node, Graph\\
URL & MVGRL~\cite{hassani2020contrastive} & GNN & Contrastive Learning &Node, Graph\\

\midrule

Fine-tuning & GraphCL~\cite{you2020graph}& GNN & Contrastive Learning &  Node, Graph\\
Fine-tuning & CSSL~\cite{zeng2021contrastive}& GNN & Contrastive Learning & Graph\\
URL\&Fine-tuning & GCC~\cite{qiu2020gcc}& GNN & Contrastive Learning &  Node, Graph\\
Fine-tuning & G-BERT~\cite{shang2019pre}& BERT &  Graph Generation & Edge\\
Fine-tuning & AdapterGNN~\cite{li2023adaptergnn}& GNN & Multi-task &  Graph\\

Fine-tuning & GROVER~\cite{rong2020self}& Graph Transformer & Property Prediction & Graph\\
URL\&Fine-tuning & Graph-Bert~\cite{zhang2020graph}& Graph Transformer & Graph Generation &  Node \\
Fine-tuning  & G-Adapter~\cite{gui2023g}& Graph Transformer & Multi-task &  Graph\\
Fine-tuning & GraphGPT~\cite{zhao2023graphgpt} & Graph Transformer & Graph Generation & Node, Edge, Graph\\

Fine-tuning & MoMu~\cite{su2022molecular} & BERT, GNN & Contrastive Learning &  Graph\\
Fine-tuning & TOUCHUP-G~\cite{zhu2023touchup} & BERT, ViT, GNN & Contrastive Learning &  Node, Edge\\

\midrule
Prompt Tuning	 & GraphPrompt~\cite{liu2023graphprompt} & GNN &  Contrastive Learning &  Node, Graph \\
Prompt Tuning & GPPT~\cite{sun2022gppt} & GNN &  Contrastive Learning & Node\\
Prompt Tuning & PGCL~\cite{gong2023prompt} & GNN & Contrastive Learning & Node, Edge, Graph\\
Prompt Tuning & GPF~\cite{fang2023universal} & GNN &  Multi-task &  Graph\\
Prompt Tuning & ProG~\cite{sun2023all} & GNN, Graph Transformer & Contrastive Learning &  Node, Edge, Graph\\
Prompt Tuning & ULTRA-DP~\cite{chen2023ultra} & GNN & Multi-task &  Node\\
Prompt Tuning & SAP~\cite{ge2023enhancing} & GNN & Contrastive Learning &  Node, Graph\\
Prompt Tuning & PRODIGY~\cite{huang2023prodigy}& GNN & Multi-task & Node, Edge\\
Prompt Tuning & SGL-PT~\cite{zhu2023sgl}& GNN & Multi-task &  Node, Graph\\
Prompt Tuning & DeepGPT~\cite{shirkavand2023deep} & Graph Transformer & Graph Regression &  Graph\\
\bottomrule
\end{tabular}
}
\end{table*}
\section{Deep Learning on Graphs}
\label{sec:graph foundation model}
With the rapid development of deep neural networks (DNNs), GNN techniques modeling graph structure and node attributes for representation learning have been widely explored and have become one key technology in Graph ML.
While vanilla GNNs demonstrate proficiency in various graph tasks, they still encounter several challenges such as scalability, generalization to unseen data, and limited capability in capturing complex graph structures.
To overcome these limitations, many efforts have been made to improve GNN with the self-supervised paradigm. 
Therefore, to provide a comprehensive review of these methods, in this section, we first introduce the backbone architecture, including GNN-based models and graph transformer-based models. 
After that, we explore two important aspects of self-supervised graph ML models: graph pretext tasks and downstream adaptation.
Note that a comprehensive summary of these methods is presented in Table~\ref{tab:Graph_foundation_model}.

\subsection{Backbone Architecture}
\label{sec:backbone}
As one of the most active fields in the artificial intelligence (AI) community,
various GNN methods have been proposed to solve various tasks. 
The powerful capability of these models is largely dependent on the development of their backbone architectures.
Therefore, in this subsection, we focus on two broadly used architectures: neighborhood aggregation-based models and graph transformer-based models.

\subsubsection{Neighborhood Aggregation-based Model}

\label{sec:neighborhood aggregation}
Neighborhood aggregation-based models are the most popular graph learning architectures that have been extensively studied and applied in various downstream tasks. 
These models operate based on the message-passing mechanism~\cite{gilmer2017neural}, which updates a node’s representation by aggregating the features of its neighboring nodes along with its own features. Formally, this process can be represented as:
\begin{align}
    m_u &= Aggregate(f_v, v \in \mathcal{N}_u),\\
    f_u' &= Update(m_u, f_u),
\end{align}
where, for each node $u$, a message $m_u$ is generated through the aggregation function from its neighboring nodes. Subsequently, the graph signal $f$ is updated with the message.

GCN is a typical method designed to leverage both the graph structure and the node attributes. This architecture updates node representations by aggregating neighboring features with the node's own. 
As the number of network layers increases, each captures an increasingly larger neighborhood. 
Owing to the efficiency and performance, GCN~\cite{kipf2016semi} has been widely applied by several methods such as CSSL~\cite{zeng2021contrastive} and  PRODIGY~\cite{huang2023prodigy}.
GraphSAGE~\cite{hamilton2017inductive} is another notable neighborhood aggregation-based model. 
Due to its inductive paradigms, GraphSAGE can easily generalize to unseen nodes or graphs, making it widely employed by many studies such as PinSage~\cite{ying2018graph} for inductive learning.
Additionally, several studies~\cite{sun2023all,huang2023prodigy,hou2022graphmae} incorporate Graph Attention Networks (GATs)~\cite{velivckovic2017graph} as the backbone architecture. GATs integrate attention mechanisms into GNNs, assigning variable weights to neighboring nodes,  thereby focusing on the most relevant parts of the input graph for improved node representations.
As another important model in the family of GNNs, Graph Isomorphism Network (GIN)~\cite{xu2018powerful} has also been widely used~\cite{liu2023graphprompt,qiu2020gcc,you2020graph,zhu2023sgl}, due to its powerful representation ability.
Its unique architecture guarantees the expressiveness equivalent to the Weisfeiler-Lehman isomorphism test, making it widely chosen as the backbone model for a lot of structure-intensive tasks.
Considering that real-world graphs are usually noisy, recent studies~\cite{luo2021learning,zheng2020robust} enhance GNN robustness against noisy by pruning irrelevant edges.

Although these models are widely adopted to solve graph tasks, they still suffer from some inherent limitations, such as over-smoothing and lack of generalization. In addition, the lower amount of parameters also limits the modeling capacity as the backbone model to serve multiple datasets and tasks.

\subsubsection{Graph Transformer-based Model}
\label{sec:graph transformer}
While traditional GNN models have shown remarkable performance in processing graph-structured data, they suffer from some limitations. 
A significant challenge for these models is their difficulty in handling large graphs due to their reliance on local neighborhood information and their limited capacity in capturing long-range dependencies within the graph~\cite{maron2019provably,kreuzer2021rethinking,morris2019weisfeiler}. 
To overcome these problems, inspired by the success of the transformer model in various NLP tasks, graph transformer-based models have been proposed~\cite{ying2021transformers,yun2019graph}. These models leverage the self-attention mechanism to adaptly capture both local and global graph structures, allowing the model to stack multiple layers without over-smoothing.
Due to the lower inductive bias, graph transformer-based models can learn structural patterns from data rather than solely relying on the graph structure.
Additionally, transformers have demonstrated great scaling behavior in CV and NLP, suggesting that their performance can keep improving with more data and parameters.

Graph transformer-based models have been widely applied as a backbone architecture in various tasks~\cite{zhang2020graph,rong2020self,li2021effective}. For example, Heterformer~\cite{jin2023heterformer} introduces a graph-empowered Transformers architecture by adding neighbor tokens into each language Transformer layer. 
Edgeformers~\cite{jin2023edgeformers} propose to encode text and structure inside each Transformer layer jointly.
Graph-Bert~\cite{zhang2020graph} employs a transformer to pre-train on the graph dataset with feature and edge reconstruction tasks and then fine-tunes for various downstream tasks.
Similarly, GROVER~\cite{rong2020self} introduces a self-supervised graph transformer-based model designed specifically for large-scale molecular data.
It pre-trains on extensive molecular datasets and then fine-tunes for specific downstream tasks.
GraphGPT~\cite{zhao2023graphgpt} employs a (semi-)Eulerian path to transform the graph into a sequence of tokens, and then feeds the sequence into the transformer. Specifically, it constructs a dataset-specific vocabulary such that each node can correspond to a unique node ID.

Despite graph transformer-based models that can somehow address the limitations of traditional GNNs, they also face several challenges. One of the challenges is the quadratic complexity caused by self-attention, which becomes particularly problematic for large-scale graphs. In addition, there is a risk of losing some information about the original graph structure when serializing the graph.

\subsection{Self-Supervised Learning on Graphs}
\label{sec:ssl}
To adapt GNNs to various graph tasks, many self-supervised learning methods have been proposed and extensively studied.
These approaches enable GNNs to learn graph representations from the pre-training task and transfer them to various downstream tasks, such as node classification, graph classification, and link prediction. 
Therefore, in this subsection, we will introduce graph self-supervised learning methods from pretext tasks and downstream adaptation, respectively.

\begin{figure}[tb]
 \centering
\subfigure{\label{fig:rl}\includegraphics[width=0.6\linewidth]{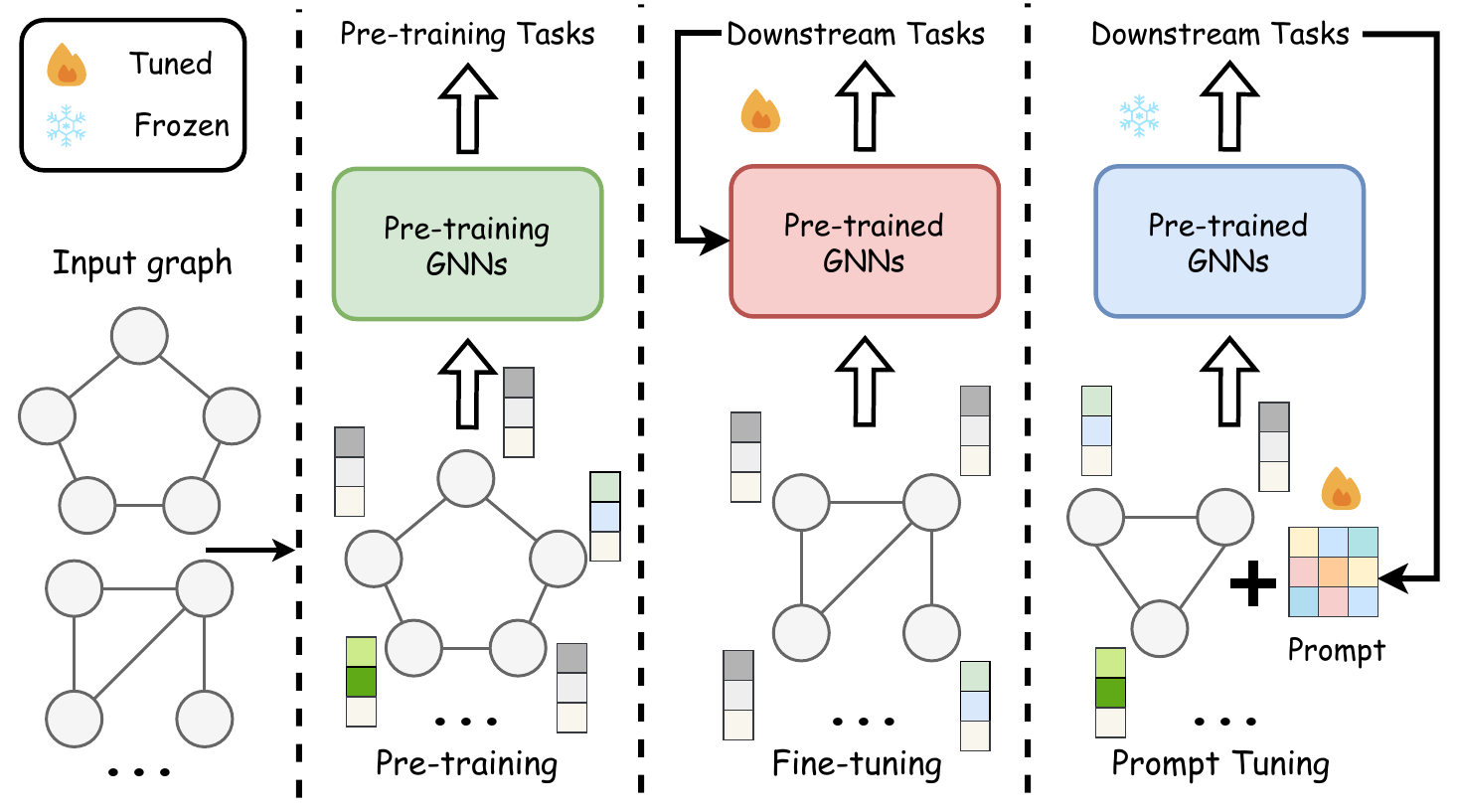}}
\vskip -0.2in
\caption{A comparison of pre-training, fine-tuning, and prompt tuning. (a) Pre-training involves training the GNN model based on specific pre-training tasks. (b) Fine-tuning updates the parameters of the pre-trained GNN model according to the downstream tasks. (c) Prompt tuning generates and updates the features of the prompt according to the downstream tasks, while keeping the pre-trained GNN model fixed and without any modification.
}
\label{fig:prompt-tuning}
\vskip -0.2in
\end{figure}

\subsubsection{Graph Pretext Tasks}
\label{sec:Graph Pretext Tasks}

\noindent \textbf{Graph Contrastive Learning} aims to learn augmentation
representations by contrasting similar and dissimilar graph data pairs,  effectively identifying nuanced relationships and structural patterns.
We can review graph contrastive learning from two perspectives: graph augmentations and the scale of contrast.

Generally, graph augmentations can be broadly categorized into two types: 1) \emph{feature perturbation} and 2) \emph{topology perturbation}. They assume that tiny changes in the feature or structural space do not change the semantics of the Node/Edge/(sub)graph.
Feature perturbation involves perturbing the features of the nodes in the graph.
For example, GRACE~\cite{zhu2020deep} randomly masks the node features to learn more robust representations.
On the other hand, topology perturbation mainly involves modifying the structure of the graph.
A typical example is CSSL~\cite{zeng2021contrastive} which employs strategies like edge perturbation or node dropping to adopt graph-graph level contrast, thereby enhancing the robustness of representations.

Regarding the scale of contrast, the approaches can be divided into node-level and graph-level. 
For example, GRACE~\cite{zhu2020deep} computes the similarities between node-level embeddings to learn discriminative node representations.
GCC~\cite{qiu2020gcc} also works at the node level but learns local structural patterns by sampling a node's neighbors to obtain subgraphs (positive pairs) and contrasting them with randomly selected non-contextual subgraphs (negative pairs).
In contrast,  DGI~\cite{velivckovic2018deep} contrasts node-level embeddings with graph-level embedding to capture global graph structures.
GraphCL~\cite{you2020graph} takes a different approach by implementing graph-to-graph level contrast, thereby learning robust representations. 
The scale used for pre-training has a huge impact on the downstream performance. When adopting contrastive learning as the pre-training task, one key challenge is how to design the objective such that the embeddings learned can account for downstream tasks of different scales.

\noindent \textbf{Graph Generation} methods aim to learn the distribution of graph data to enable graph generation or reconstruction. 
In contrast to models in CV that predict masked image patches, or in NLP that predict the next token in a sequence, graph data presents a unique challenge due to its interconnected nature.
Consequently, graph generation methods typically work on the feature or structural space.
Feature generation methods focus on masking the features of one or a subset of nodes and then training the model to recover the masked features.
For instance, GraphMAE~\cite{hou2022graphmae} utilizes a masked autoencoder framework to reconstruct masked graph portions based on their context, effectively capturing the underlying node semantics and their connection patterns. 
Alternatively, structure generation methods concentrate on training the model to recover the graph structure.
The method GraphGPT~\cite{zhao2023graphgpt} encodes the graph into sequences of tokens and then employs a transformer decoder to predict the next token of the sequence to recover the connectivity of the graph.
In addition, Graph-Bert~\cite{zhang2020graph} is trained on both node attribute recovery and graph structure recovery tasks to ensure that the model captures local node attribute information while maintaining a global view of the graph structure.

\noindent \textbf{Graph Property Prediction}
methods gain guidance from the node-, edge-, and graph-level properties, which are inherently present in the graph data. These methods follow a training approach similar to supervised learning, as both utilize "sample-label" pairs for training. The key distinction lies in the origin of the labels: in supervised learning, labels are manually annotated by human experts which can be costly in real scenarios, whereas in property-based learning, the labels are automatically generated from the graph using some heuristics or algorithms.
For example, GROVER~\cite{rong2020self} utilizes professional software
to extract the information on graph motifs as labels for classification.
Similarly, ~\cite{hu2019pre} leverages statistical properties of the graph for graph self-supervised learning.

\subsubsection{Downstream Adaptation}
\label{sec:Downstream Adaptation}
\noindent \textbf{Unsupervised Representation Learning} (URL) is a common method due to the scarcity of labeled data in the real world~\cite{velivckovic2018deep,zhu2020deep,hou2022graphmae,hassani2020contrastive}.
In URL, the pre-trained graph encoder is frozen and only a task-specific layer is learned during downstream tuning.
The learned representations are then directly fed into decoders. 
This pattern allows URLs to be efficiently applied to downstream tasks.
For example, DGI~\cite{velivckovic2018deep} trains an encoder model to learn node representations within graph-structured.
Node representations can then be used for downstream tasks.
However, due to the gap between the pretext task and downstream tasks, URL can also lead to suboptimal performance.

\noindent \textbf{Fine-tuning}
is the default method to adapt a pre-trained model to a certain downstream task. As shown in Figure~\ref{fig:prompt-tuning}, it adds a randomly initialized task header (e.g., a classifier) on top of the pre-trained model, and during fine-tuning, both the backbone model and the header are jointly trained~\cite{zhang2020graph,you2020graph,zeng2021contrastive}. 
Compared with URL, fine-tuning provides more flexibility as it allows changes in the backbone parameters, and one can choose the layers to be tuned while keeping others fixed.
Additionally, recent studies~\cite{li2023adaptergnn,gui2023g,you2020graph} further explore advanced graph fine-tuning methods that go beyond naive fine-tuning.
For instance, AdapterGNN~\cite{li2023adaptergnn} introduces two trainable adapters in parallel before and after the message passing. It freezes the GNN model during fine-tuning while only tuning the adapters, enabling parameter-efficient fine-tuning with minimal influence on the downstream performance.

\noindent \textbf{Prompt-tuning:}
"Pre-training \& fine-tuning" is prevalent in adapting pre-trained models to specific downstream tasks, but it overlooks the gap between pre-training and downstream tasks, potentially limiting generalization capabilities. Moreover, fine-tuning for different tasks also leads to significant time and computational costs.
Inspired by recent advancements in NLP, several methods~\cite{liu2023graphprompt,zhu2023sgl,sun2022gppt,sun2023all,gong2023prompt,fang2023universal,ge2023enhancing,chen2023ultra,shirkavand2023deep} have presented the potential of introducing prompts to adapt pre-trained models to specific tasks as illustrated in Figure~\ref{fig:prompt-tuning}. 
Specifically, Prompt-tuning first unifies the downstream task with the pre-trained task into the same paradigm, followed by the introduction of learnable prompts for tuning.
For example, GPPT~\cite{sun2022gppt} first reframe node classification as link predictions. 
GraphPrompt~\cite{liu2023graphprompt} further extends graph classification into link prediction.
On the other hand, Prog~\cite{sun2023all} unifies all the downstream tasks into subgraph classification. 
The inserting prompt including vectors~\cite{sun2022gppt,liu2023graphprompt,fang2023universal}, node~\cite{zhu2023sgl} and sub-graph~\cite{sun2023all}.
By inserting these prompts, the pre-trained parameters can be utilized in a way that aligns more closely with the requirements of the downstream tasks.

\begin{figure*}[tb]
    \centering
    \includegraphics[width=0.9\linewidth]{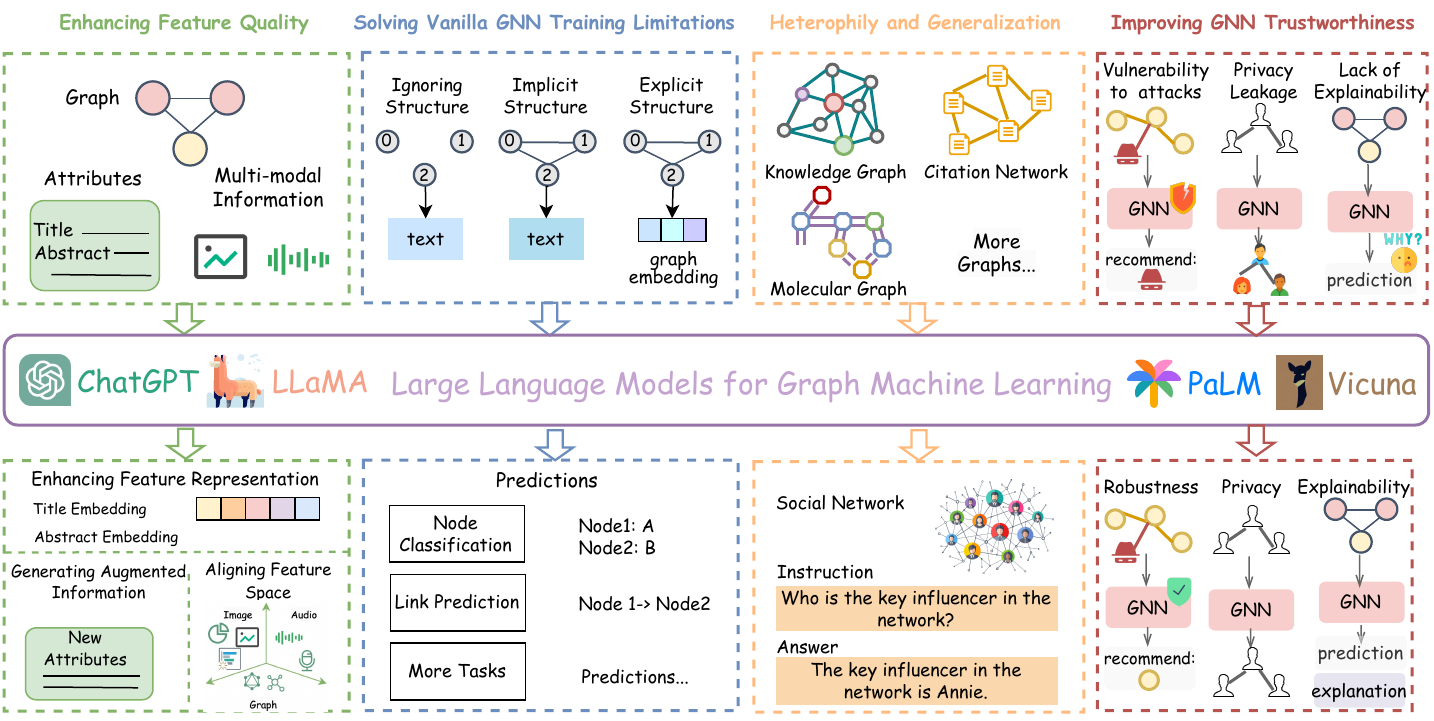}
    \caption{Illustration of LLMs for Graph ML.
    (1) Methods using LLMs for \textit{Enhancing Feature Quality} by enhancing feature representation, generating augmented information, and aligning feature space. 
    (2) Explorations for \textit{Solving Vanilla GNN Training Limitations} are categorized based on how structural information in the graph is processed: ignoring structural information, implicit structural information, and explicit structural information. 
    (3) Research about employing LLMs to alleviate the limitations of \textit{Heterophily and Generalization}.
    (4) Studies leveraging LLMs for \textit{Improving GNN Trustworthiness}.}
    \label{fig:llm-enhanced-graph}
\vskip -0.2in
\end{figure*}

\section{LLMs for Graph Models}\label{sec:LLM-enhanced Graph}

Despite great potential, Graph ML based on the GNNs has its inherent limitations. 
Firstly, vanilla GNN models commonly demand labeled data for supervision, and obtaining these annotations can be resource-intensive in terms of time and cost. Secondly, real-world graphs often contain abundant textual information, which is crucial for downstream tasks. However, GNNs typically rely on shallow text embeddings for semantic extraction, thereby limiting their capacity to capture intricate semantics and text features. 
Moreover, the diversity of graphs presents challenges for GNN models in terms of generalization across diverse domains and tasks.

Recently, LLMs have achieved remarkable success in handling natural language, with exciting features like (1) conducting zero/few-shot predictions and (2) providing a unified feature space. These capabilities present a potential solution to address the above challenges faced by Graph ML.
Therefore, this section aims to investigate the contributions that current LLMs can make to enhance Graph ML, while also examining their current limitations, as Figure~\ref{fig:llm-enhanced-graph} shows.

\begin{table*}[t]
  \centering
  \caption{A summary of LLM for Graph ML research. We present the GNN model, LLM model, predictor, domain, task, datasets, and project link. \textbf{FT} is Fine-tuning, refers to whether modifications are made to the parameters of the LLM model while \textbf{PR} is Prompting, involves inputting textual prompts to the LLM to obtain responses. In the context of \textbf{task}, \textcolor{blue}{\ding{108}} denotes node-level tasks such as node classification, \textcolor{green}{\ding{108}} signifies edge-level tasks like link prediction, \textcolor{orange}{\ding{108}} represents graph-level tasks such as graph classification, and \textcolor{myorange}{\ding{108}} pertains to structure understanding tasks, such as node degree counting.}
  \resizebox{\textwidth}{!}{
    \begin{tabular}{cccccccccccc}
    \toprule
    Role & Sub Category & Method & GNN Model & LLM Model & FT & PR & Domain & Task & \multicolumn{1}{l}{Datasets} & Link \\
    \midrule

\multicolumn{1}{c}{\multirow{15}[8]{*}{\rotatebox{90}{Enhancing Feature Quality}}} & \multicolumn{1}{c}{\multirow{4}[2]{*}{\makecell[c]{Enhancing Feature\\Representation}}} 

& Chen et al.~\cite{chen2023exploring} & \makecell[c]{GCN, GAT, MLP} & ChatGPT, LLaMA & \red{$\times$} & \red{$\times$} & Citation, E-commerce  & \textcolor{blue}{\ding{108}} & 5 & \href{https://github.com/CurryTang/Graph-LLM}{link} \\
      &   & SimTeG~\cite{duan2023simteg} & \makecell[c]{SAGE, MLP, etc.} & \makecell[c]{all- MiniLM-L6-v2, etc.} & \lightgreen{$\checkmark$} & \red{$\times$} & Citation, E-commerce & \textcolor{blue}{\ding{108}} \textcolor{green}{\ding{108}} & 3 & \href{https://github.com/vermouthdky/SimTeG}{link} \\
      &   & LKPNR~\cite{hao2023lkpnr} & wiki KG & \makecell[c]{ChatGLM2, RWKV,\\LLaMA2} & \lightgreen{$\checkmark$} & \red{$\times$} & Recommendation & \textcolor{myorange}{\ding{108}} & 1 & \href{https://github.com/Xuan-ZW/LKPNR}{link} \\
      &   & GRID~\cite{ni2023grid} & GAT & INSTRUCTOR & \red{$\times$} & \red{$\times$} & Robotics & \textcolor{myorange}{\ding{108}} & 2 & \href{https://jackyzengl.github.io/GRID.github.io/}{link} \\

\cmidrule{2-11}      & \multicolumn{1}{c}{\multirow{9}[2]{*}{\makecell[c]{Generating Augmented\\Information}}} 
      & TAPE~\cite{he2023harnessing} & \makecell[c]{GCN, SAGE,MLP,\\RevGAT} & ChatGPT, LLaMA2 & \red{$\times$} & \lightgreen{$\checkmark$} & Citation, E-commerce & \textcolor{blue}{\ding{108}} & 5 & \href{https://github.com/XiaoxinHe/TAPE}{link} \\
      &   & KEA~\cite{chen2023exploring} & \makecell[c]{GCN, GAT, MLP} & ChatGPT & \red{$\times$} & \lightgreen{$\checkmark$} & Citation, E-commerce & \textcolor{blue}{\ding{108}} & 5 & \href{https://github.com/CurryTang/Graph-LLM}{link} \\
      &   & RLMRec~\cite{ren2023representation} & LightGCN & ChatGPT & \red{$\times$} & \lightgreen{$\checkmark$} & Recommendation & \textcolor{green}{\ding{108}} & 3 & \href{https://github.com/HKUDS/RLMRec}{link} \\
      &   & LLMRec~\cite{weillmrec} & LightGCN & \makecell[c]{ChatGPT} & \red{$\times$} & \lightgreen{$\checkmark$} & Recommendation & \textcolor{blue}{\ding{108}} & 2 & \href{https://github.com/HKUDS/LLMRec}{link} \\
      &   & LLM-Rec~\cite{lyu2023llmrec} & - & text-davinci-003 & \lightgreen{$\checkmark$} & \red{$\times$} & Recommendation & \textcolor{green}{\ding{108}} & 2 & - \\
      &   & LLM4Mo~\cite{qian2023can} & - & ChatGPT & \red{$\times$} & \lightgreen{$\checkmark$} & Molecular & \textcolor{orange}{\ding{108}} & 3 & \href{https://github.com/ChnQ/LLM4Mol}{link} \\
      &   & GPT-MolBERTa~\cite{balaji2023gptmolberta} & - & ChatGPT & \red{$\times$} & \lightgreen{$\checkmark$} & Molecular & \textcolor{orange}{\ding{108}} & 9 & - \\
      &   & ENG~\cite{yu2023empower} & GCN,GAT & ChatGPT & \red{$\times$} & \lightgreen{$\checkmark$} & Citation & \textcolor{blue}{\ding{108}} & \multicolumn{1}{c}{-} &  \\
      &   & Sun et al.~\cite{sun2023large} & GAT, GCN, etc. & ChatGPT & \red{$\times$} & \lightgreen{$\checkmark$} & Citation & \textcolor{blue}{\ding{108}} & 4 & - \\

\cmidrule{2-11}      & \multicolumn{1}{c}{Aligning Feature Space} 
    & TouchUp-G~\cite{zhu2023touchup} & SAGE, GAT, etc. & BERT, etc. & \lightgreen{$\checkmark$} & \red{$\times$} & Citation, E-commerce & \textcolor{blue}{\ding{108}} & 4 & \href{https://github.com/jwzhi/TouchUp-G}{link} \\
    &   & OFA~\cite{liu2023one} & GCN,GAT,etc. & LLaMA-2-7B, etc. & \lightgreen{$\checkmark$} & \red{$\times$} & Citation, Molecular, etc. & \textcolor{blue}{\ding{108}} \textcolor{green}{\ding{108}} \textcolor{orange}{\ding{108}} & 9 & \href{https://github.com/LechengKong/OneForAll}{link} \\

    \midrule

    \multicolumn{1}{c}{\multirow{25}[6]{*}{\rotatebox{90}{Solving Vanilla GNN Training Limitations}}} & \multicolumn{1}{c}{\makecell[c]{Ignoring Structural\\Information}} &  Hu et al.~\cite{hu2023text} &  - & ChatGPT,GPT-4 & \red{$\times$} & \lightgreen{$\checkmark$} & Citation,KG & \textcolor{blue}{\ding{108}} \textcolor{green}{\ding{108}} & 5 & - \\
    
\cmidrule{2-11}      & \multicolumn{1}{c}{\multirow{14}[2]{*}{\makecell[c]{Implicit Structural\\Information}}} & 

    GPT4Graph~\cite{guo2023gpt4graph} & - & text-davinci-003 & \red{$\times$} & \lightgreen{$\checkmark$} & - & \textcolor{myorange}{\ding{108}}\textcolor{orange}{\ding{108}}\textcolor{blue}{\ding{108}} & 4 & \href{https://anonymous.4open. science/r/GPT4Graph}{link} \\
      &   & GraphText~\cite{zhao2023graphtext} & - & ChatGPT, LLaMA2-7B & \lightgreen{$\checkmark$} & \lightgreen{$\checkmark$} & Citation, Web & \textcolor{blue}{\ding{108}} & 7 & - \\
      
      &   & NLGraph~\cite{wang2023can} & - & ChatGPT, GPT-4, etc. & \red{$\times$} & \lightgreen{$\checkmark$} & - & \textcolor{myorange}{\ding{108}} & 3 & \href{https://github.com/Arthur-Heng/NLGraph}{link} \\
      &   & InstructGLM~\cite{ye2023natural} & - & Flan T5, LLaMA & \lightgreen{$\checkmark$} & \lightgreen{$\checkmark$} & Citation & \textcolor{blue}{\ding{108}} & 3 & \href{https://github.com/Graphlet-AI/llm-graph-ai}{link} \\
      &   & LLMtoGraph~\cite{liu2023evaluating} & - & \makecell[c]{ChatGPT, GPT-4,\\Vicuna-13B, etc.} & \red{$\times$} & \lightgreen{$\checkmark$} & - & \textcolor{myorange}{\ding{108}} & - & \href{https://github.com/Ayame1006/LLMtoGraph}{link} \\
      &   & Graph Agent~\cite{wang2023graph} & - & \makecell[c]{GPT-4,\\embedding-ada-002} &  \red{$\times$} & \lightgreen{$\checkmark$} & Citation, Bioinformatics & \textcolor{blue}{\ding{108}} \textcolor{green}{\ding{108}} & 2 & - \\
      &   & LLM-Prop~\cite{rubungo2023llmprop} & - & T5 & \lightgreen{$\checkmark$} & \red{$\times$} & Material Science & \textcolor{orange}{\ding{108}} & \multicolumn{1}{c}{-} & \href{https://github.com/vertaix/LLM-Prop}{link} \\
      &   & GLRec~\cite{wu2023exploring} & - & BELLE-LLaMA-7B & \lightgreen{$\checkmark$} & \lightgreen{$\checkmark$} & Recommendation & \textcolor{green}{\ding{108}} & 1 & - \\
      &   & ReLM~\cite{shi2023relm} & TAG,GCN & GPT-3.5, Vicuna & \red{$\times$} & \lightgreen{$\checkmark$} & Chemistry & \textcolor{orange}{\ding{108}} & 5 & \href{https://github.com/syr-cn/ReLM}{link} \\
      &   & Chen et al.~\cite{chen2023exploring} & - & ChatGPT & \red{$\times$} & \lightgreen{$\checkmark$} & Citation, E-commerce & \textcolor{blue}{\ding{108}} & 5 & \href{https://github.com/CurryTang/Graph-LLM}{link} \\
      &   & Hu et al.~\cite{hu2023text} &   & ChatGPT,GPT-4 & \red{$\times$} & \lightgreen{$\checkmark$} & Citation,KG & \textcolor{blue}{\ding{108}} \textcolor{green}{\ding{108}} & 5 & - \\
      &   & Huang et al.~\cite{huang2023can} & - & ChatGPT & \red{$\times$} & \lightgreen{$\checkmark$} & Citation, E-commerce & \textcolor{blue}{\ding{108}} & 5 & \href{https://github.com/TRAIS-Lab/LLM-Structured-Data}{link} \\
      &   & Fatemi et al.~\cite{fatemi2023talk} &   & PaLM2 XXS, PaLM 62B & \red{$\times$} & \lightgreen{$\checkmark$} & - & \textcolor{myorange}{\ding{108}} & \multicolumn{1}{c}{-} & - \\
      &   & MolReGPT~\cite{li2023empowering} & - & ChatGPT & \red{$\times$} & \lightgreen{$\checkmark$} & Molecular & \textcolor{orange}{\ding{108}} & 1 & \href{https://github.com/phenixace/MolReGPT}{link} \\
      &   & LLaGA~\cite{chen2024llaga} & - & Vicuna-7B & \red{$\times$} & \lightgreen{$\checkmark$} & Citation, E-commerce & \textcolor{blue}{\ding{108}} \textcolor{green}{\ding{108}} & 4 & \href{https://github.com/VITA-Group/LLaGA}{link} \\
       &   & GraphEdit~\cite{guo2024graphedit} & GCN & Vicuna-v1.5 & \lightgreen{$\checkmark$} & \lightgreen{$\checkmark$} & Citation & \textcolor{blue}{\ding{108}} & 3 & \href{https://github.com/HKUDS/GraphEdit}{link} \\

\cmidrule{2-11}      & \multicolumn{1}{c}{\multirow{10}[2]{*}{\makecell[c]{Explicit Structure\\Information}}} & 

GraphGPT~\cite{tang2023graphgpt} & Graph Transformer & Vicuna-7B & \lightgreen{$\checkmark$} & \lightgreen{$\checkmark$} & Citation & \textcolor{blue}{\ding{108}} & 3 & \href{https://graphgpt.github.io/c}{link} \\
      &   & GraphLLM~\cite{chai2023graphllm} & Graph Transformer & LLaMA2 & \red{$\times$} & \lightgreen{$\checkmark$} & - & \textcolor{myorange}{\ding{108}} & 4 & \href{https://github.com/mistyreed63849/Graph-LLM}{link} \\
      &   & GNP~\cite{tian2023graph} & GAT & Flan T5 & \lightgreen{$\checkmark$} & \lightgreen{$\checkmark$} & KG & \textcolor{orange}{\ding{108}} & 4 & - \\
      &   & DrugChat~\cite{liang2023drugchat} & GAT, etc. & Vicuna-13B & \red{$\times$} & \lightgreen{$\checkmark$} & Drug & \textcolor{orange}{\ding{108}} & \multicolumn{1}{c}{-} & \href{https://github.com/UCSD-AI4H/drugchat}{link} \\
      &   & KoPA~\cite{zhang2023making} & RotateE & Alpaca-7B & \lightgreen{$\checkmark$} & \lightgreen{$\checkmark$} & Knowledge Graph & \textcolor{green}{\ding{108}} & 3 & \href{https://github.com/zjukg/KoPA }{link} \\
      &   & GIMLET~\cite{zhao2023gimlet} & - & T5 & \lightgreen{$\checkmark$} & \lightgreen{$\checkmark$} & Molecular & \textcolor{orange}{\ding{108}} & 14 & \href{https://github.com/zhao-ht/GIMLET}{link} \\
      &   & GIT-Mol~\cite{liu2023gitmol} & GIN & MolT5 & \lightgreen{$\checkmark$} & \lightgreen{$\checkmark$} & Molecular & \textcolor{orange}{\ding{108}} & 6 & - \\
      &   & BioMedGPT~\cite{luo2023biomedgpt} & GIN & LLaMA2-7B-Chat & \lightgreen{$\checkmark$} & \lightgreen{$\checkmark$} & Biomedical & \textcolor{orange}{\ding{108}} & 3 & \href{https://github.com/PharMolix/OpenBioMed}{link} \\
      &   & ProteinChat~\cite{guo2023proteinchat} & GVP-GNN & Vicuna-13B & \red{$\times$} & \lightgreen{$\checkmark$} & Protein & \textcolor{orange}{\ding{108}} & 1 & \href{https://github.com/UCSD-AI4H/proteinchat}{link} \\
      &   & DGTL~\cite{qin2023disentangled} & Disentangled GNN & LLaMA-2-13B-chat & \red{$\times$} & \lightgreen{$\checkmark$} & Citation, E-commerce & \textcolor{blue}{\ding{108}} & 3 & - \\

      &   & G-Retriever~\cite{he2024g} & GAT & LLaMA-2-7B & \lightgreen{$\checkmark$} & \lightgreen{$\checkmark$} & - & \textcolor{orange}{\ding{108}} & 3 & \href{https://github.com/XiaoxinHe/G-Retriever}{link} \\

      & & GraphToken~\cite{perozzi2024let} & GCN,GIN,etc. & PaLM 2 S & \red{$\times$} & \lightgreen{$\checkmark$} & - & \textcolor{myorange}{\ding{108}} & 1 & - \\
    \midrule

    \multicolumn{1}{r}{\multirow{2}[2]{*}{\rotatebox{90}{HG}}} 
    &  Heterophily & Chen et al.~\cite{chen2023exploring} & - & ChatGPT & \red{$\times$} & \lightgreen{$\checkmark$} & Citation, E-commerce & \textcolor{blue}{\ding{108}} & 5 & \href{https://github.com/CurryTang/Graph-LLM}{link} \\
      & \multirow{2}[2]{*}{Generalization}  & GraphText~\cite{zhao2023graphtext} & - & ChatGPT, LLaMA2-7B & \lightgreen{$\checkmark$} & \lightgreen{$\checkmark$} & Citation, Web & \textcolor{blue}{\ding{108}} & 7 & - \\
          &   & OpenGraph~\cite{xia2024opengraph} & Graph Transformer & Not mentioned &  \red{$\times$} & \lightgreen{$\checkmark$} & Citation, Drug, etc. & \textcolor{blue}{\ding{108}} \textcolor{green}{\ding{108}} & 7 & \href{https://github.com/HKUDS/OpenGraph}{link} \\
    \midrule
    
    \multicolumn{1}{r}{\multirow{3}[2]{*}{\rotatebox{90}{Trustworthy}}} 
    &  \multirow{1}[2]{*}{Robustness} & \citet{guo2024learning} & GCN, etc. & GPT-3.5, etc. & \lightgreen{$\checkmark$} & \lightgreen{$\checkmark$} & Citation & \textcolor{blue}{\ding{108}} & 6 & \href{https://github.com/KaiGuo20/GraphLLM_Robustness}{link} \\
    
      &   & 
      LLM4RGNN~\cite{zhang2024can} & GCN, etc. & Llama3-8B, etc. & \lightgreen{$\checkmark$} & \lightgreen{$\checkmark$} & Citation, E-commerce & \textcolor{blue}{\ding{108}} & 5 & \href{https://github.com/zhongjian-zhang/LLM4RGNN}{link} \\

      & {Privacy}  & ~\citet{guan2024large} & GCN & Vicuna-7B & \lightgreen{$\checkmark$} & \lightgreen{$\checkmark$} & Citation & \textcolor{green}{\ding{108}} & 4 & - \\
         
    & {Explainability}  & LLM-GCE~\cite{he2024explaining} & GCN & GPT-4, GPT-3.5 & \red{$\times$} & \lightgreen{$\checkmark$} & Medical & \textcolor{blue}{\ding{108}} & 5 & \href{https://github.com/YinhanHe123/new_LLM4GNNExplanation}{link} \\
    
    \bottomrule
    \end{tabular}
    }
  \label{tab:llm-enhanced graph}%
  \vskip -0.2in
\end{table*}%

\subsection{Enhancing Feature Quality}
\label{sec:enhance feature}
Graphs encompass diverse attribute information, spanning text, images, audio, and other multi-modal modes. The semantics of these attributes play a crucial role in a range of downstream tasks. In comparison with earlier pre-trained models, LLMs stand out due to their substantial parameter volume and training on extensive datasets, endowing them with rich open-world knowledge. Consequently, researchers are exploring the potential of LLMs to improve feature quality and align feature space. This section delves into research endeavors aimed at leveraging LLMs to accomplish these goals.

\subsubsection{Enhancing Feature Representation}
\label{sec:enhance rep}
Researchers utilize the powerful language understanding capabilities of LLMs to generate better representations for text attributes compared to traditional shallow text embeddings~\cite{chen2023exploring,hao2023lkpnr,ni2023grid}. For example, Patton~\cite{jin2023patton} proposes to pre-train a language model on the target graph to obtain high-quality feature representation.
METERN~\cite{jin2023learning} introduces a soft prompt-based method to learn node multiplex embeddings for different edge types with one language model encoder.
Chen et al.~\cite{chen2023exploring} utilize LLMs as text encoders and the GNN model as a predictor, validating the effectiveness of LLMs as an enhancer in node classification tasks. In LKPNR~\cite{hao2023lkpnr}, an LK-Aug news encoder enhances the news recommender system by concatenating LLM embeddings with entity embeddings within the news text to obtain an enriched news representation. Several researchers explore fine-tuning LLMs to obtain text representations better suited for downstream graph tasks. SimTeG~\cite{duan2023simteg} treats node classification and link prediction tasks as text classification and text similarity tasks, fine-tuning PLMs using LoRA ~\cite{hu2021lora} on the TAG dataset. The fine-tuned PLMs are then used to generate embeddings for text attributes, followed by GNN training for downstream tasks.

\subsubsection{Generating Augmented Information}
\label{sec:Generating Augmented Information}
Several studies investigate leveraging the generation capabilities and general knowledge of LLMs to generate augmented information from original textual attributes. This augmented information can enhance the quality and depth of features available for graph learning, leading to improved model performance and generalization. TAPE~\cite{he2023harnessing} utilizes LLMs to generate potential node labels and explanations from textual attributes, such as titles and abstracts. These generated labels and explanations are treated as augmented attributes. A fine-tuned language model (LM) then encodes these attributes, which are subsequently processed by a GNN model to incorporate the graph structure for final predictions. In contrast to TAPE, KEA~\cite{chen2023exploring} does not directly use LLM-generated node labels as augmented information for graphs. 
Instead, LLM is employed to extract terms mentioned in textual attributes and provide detailed descriptions of these terms. In the domain of molecular graphs, LLM4Mol~\cite{qian2023can} and GPT-MolBERTa~\cite{balaji2023gptmolberta} utilize LLMs to interpret Simplified Molecular-Input Line-Entry System (SMILES) notations. These interpretations are considered augmented attributes to enhance the representation of graph structures. In recommender systems, researchers also explore leveraging LLMs to enrich the textual attributes of user-item interaction graphs. LLM-Rec~\cite{lyu2023llmrec} prompts LLMs to generate detailed descriptions for item nodes by explicitly stating the recommendation intent. Meanwhile, LLMRec~\cite{weillmrec} produces augmented information for user nodes by utilizing historical behavior data. It outputs user profiles, including attributes like age, gender, country, language, and preferred or disliked genres as auxiliary information. This auxiliary information enhances the GNN model's understanding of user-item interaction graphs, ultimately improving recommendation performance.

In addition to generating augmented text attributes, researchers also employ LLMs to enhance graph topological structures by generating auxiliary nodes or refining existing nodes and edges. In ENG~\cite{yu2023empower}, LLM is employed to generate new nodes and their corresponding text attributes for each node category. To integrate the generated nodes into the original graph, the authors train an edge predictor using relations in the original dataset as supervised signals. Sun et al.~\cite{sun2023large} leverage LLMs to refine graph structures. Specifically, they let LLMs remove unreliable edges by predicting the semantic similarity between node attributes. Additionally, they utilize pseudo-labels generated by LLMs to aid the GNN in learning proper edge weights.

\subsubsection{Aligning Feature Space}
\label{sec:Aligning Feature Space}
In real-world scenarios, the text attributes of graphs across different domains exhibit considerable diversity. Additionally, beyond text modal attributes, the graph may encompass various other modal attributes. Employing Pretrained Models (PMs) directly for encoding cross-domain and multi-modal features may not produce satisfactory results. Therefore, LLMs are employed to align feature space and provide better representations. TouchUp-G~\cite{zhu2023touchup} introduces a graph-centric fine-tuning strategy aimed at enhancing multi-modal features for graph-related tasks. Initially, they present a novel feature homophily measure to quantify the alignment between node features and the graph structure. Building upon this measure, the authors devise a structure-aware loss function to optimize the PM by minimizing discrepancies between features and graphs. The work of~\cite{liu2023one} introduces OFA, a unified framework for classification tasks in graphs across different domains. OFA collects nine text-attributed graph datasets covering diverse domains and represents nodes and relations in natural language. LLMs are then employed to embed those cross-domain graph information into the same embedding space. Moreover, OFA proposes a graph prompting paradigm, which incorporates a prompt graph containing downstream task information into the original input graph, allowing the GNN model to adaptively perform different tasks based on the prompt graph.

\subsection{Solving Vanilla GNN Training Limitations}
\label{sec:Solving Vanilla GNN Training Limitations}
The training of vanilla GNNs relies on labeled data. However, obtaining high-quality labeled data has long been associated with substantial time and costs. In contrast to GNNs, LLMs showcase robust zero/few-shot capabilities and possess expansive open-world knowledge. This unique characteristic empowers LLMs to directly leverage node information for prediction, without relying on extensive annotated data. Therefore, researchers have explored employing LLMs to generate annotations or predictions, alleviating dependence on human supervision signals in Graph ML. According to how structural information in graph data is processed, we categorize the methods into the following three categories:

\begin{itemize}
        \item Ignoring structural information: utilize node attributes exclusively for constructing textual prompts, disregarding neighboring labels and relations.
        \item Implicit Structural information: describe neighbor information and graph topology structure in natural language;
        \item Explicit Structural information: employ GNN models to encode graph structure.
\end{itemize}

\begin{figure*}[tb]
 \centering
 \vskip -0.1in
\includegraphics[width=0.9\linewidth]{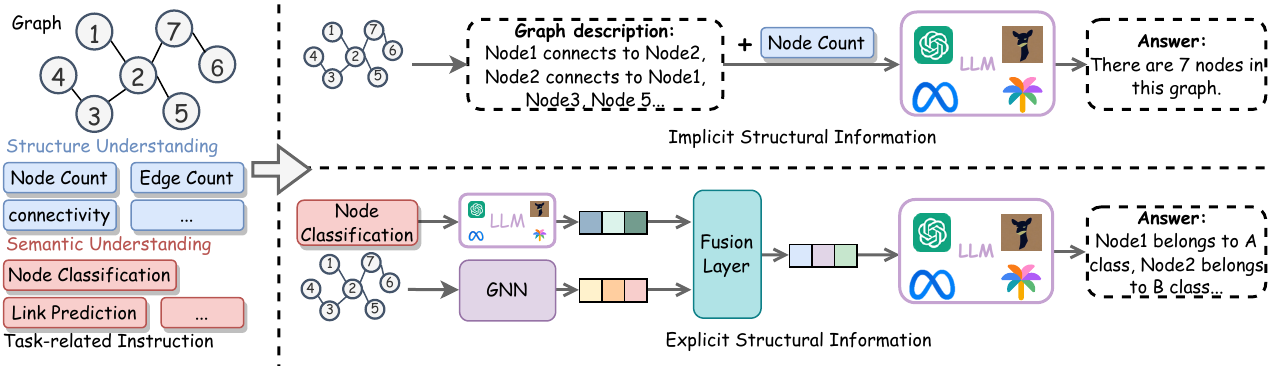}
\vskip -0.1in
\caption{The illustration of employing LLMs with implicit and explicit structural information. 
(1) Methods leveraging \textit{implicit structural information} describe nodes and graph structure information in natural language and combine task-specific instructions to form a textual prompt, which is then input into the LLM to generate prediction results. 
(2) Methods employing \textit{explicit structural information} use GNNs and LLMs to encode graph and instruction information separately. Then, fusion layers are added to align the graph and text modalities, and the fused embedding is input into the LLM for prediction. } 
\label{fig:llm as predictor}
\vskip -0.2in
\end{figure*}

The first category disregards structural information entirely, while the latter two incorporate it in different ways, as depicted in Figure~\ref{fig:llm as predictor}. Below, we introduce the relevant research for each approach.

\subsubsection{Ignoring Structural Information}

\label{sec:Ignoring Structural Information}
The fundamental distinction between graphs and text lies in the structural information inherent in graphs. Given that the LLM processes text as its input, an intuitive approach involves leveraging the textual attributes of the target node, disregarding the structural information within the graph, and making predictions directly. For instance, the work of ~\cite{hu2023text} explores the effectiveness of LLMs in solving graph tasks without using structure information. In the citation network, they employ the article's title and abstract to construct a prompt and instruct the LLM to predict the article's category. Since this kind of paradigm does not incorporate the structural information of the graph, the actual task performed by the LLM is text classification rather than a graph-related task.

\subsubsection{Implicit Structural Information}
\label{sec:Implicit Structural Information}

Researchers implicitly leverage structural information to solve graph tasks by describing graph structure in natural language. For example, Hu et al.~\cite{hu2023text} propose two kinds of methods for utilizing structural information. The first method involves directly inputting the data of all neighboring nodes into LLM, while the second method employs a retrieval-based prompt to guide the LLM to focus solely on relevant neighbor data. Similarly, Huang et al.~\cite{huang2023can} employ an LLM to assign scores to neighboring nodes and subsequently choose high-scoring nodes as structural information. NLGraph~\cite{wang2023can} introduces a Build-a-Graph prompting strategy to improve the LLM's understanding of graph structure. This strategy entails appending ``Let’s construct a graph with the nodes and edges first." after providing the graph data description. The work of~\cite{ye2023natural} introduces InstructGLM, which utilizes natural language for graph description and fine-tunes Flan-T5 through instruction tuning. They generate a set of 31 prompts by combining four configuration parameters: task type, inclusion of node features, maximum hop order, and utilization of node connections. Notably, maximum hop order and node connections implicitly convey graph structure information to the LLM. GraphEdit~\cite{guo2024graphedit} leverages LLMs to understand graph structure and refine it by removing noisy edges and uncovering implicit node connections. Specifically, it employs an edge predictor to identify the top k candidate edges for each node, and these candidate edges, along with the original edges of the graph, are then fed into the LLM. The LLM is prompted to determine which edges should be integrated into the final graph structure.

In addition to employing natural language expression, several researchers leverage structured languages for graph description. GPT4Graph~\cite{guo2023gpt4graph}, for instance, utilizes Graph Modelling Language~\cite{himsolt1997gml} and Graph Markup Language~\cite{brandes2013graph} to represent graph structure in XML format. GraphText~\cite{zhao2023graphtext} constructs a graph syntax tree for each graph, containing node attributes and relations information. By traversing this tree, structural graph-text sequences can be generated. The advantage of GraphText lies in the ability to integrate the typical inductive bias of GNNs through the construction of various graph syntax trees.

\subsubsection{Explicit Structural Information}
\label{sec:Explicit Structural Information}
While implicitly describing structure in natural language has achieved preliminary success, these methods still face certain limitations. 
Firstly, due to the constraint of input length, LLMs can only get local structural information, and lengthy contexts might diminish their reasoning~\cite{liu2023lost} and instruction-following abilities~\cite{chen2023exploring}.
Secondly, for different tasks and datasets, substantial effort is often required for prompt engineering. A prompt that performs well on one dataset may not generalize effectively to others, resulting in a lack of robustness. 
Consequently, researchers investigate representing graph structure explicitly, typically comprising three essential modules:  \emph{encoding module, fusion module}, and \emph{LLM module}. 
More specifically, the encoding module aims to process the graph-structured and textual information, generating graph embeddings and text embeddings, respectively. 
Afterward, the fusion module takes these two embeddings as input, producing a modality fusion embedding. 
At last, the modality fusion embedding, which contains both graph information and instruction information, is fed into the LLM to obtain the final answer.
Given the research focus is on how LLMs explicitly utilize graph structure information, we will delve into the encoding and fusion modules of various studies in detail, without primarily focusing on the LLM model itself.

\noindent \textbf{Encoding Module.} 
The encoding module is responsible for both graph and text encoding, and we will provide separate summaries for each.

\begin{itemize} [leftmargin=*]
    \item \textit{Graph Encoding.} Pre-trained GNN models are commonly used for graph encoding. For instance, GIT-Mol~\cite{liu2023gitmol} employs the GIN model from the pre-trained MoMu model~\cite{su2022molecular} to encode molecular graphs. KoPA~\cite{zhang2023making} utilizes the pre-trained RotateE model to obtain embeddings for entities and relations in the knowledge graph. 
    In addition, GIMLET~\cite{zhao2023gimlet} presents a unified graph-text model without the need for additional graph encoding modules. Particularly, GIMLET proposes a distance-based joint position embedding method, where the shortest graph distance is utilized to represent the relative positions between graph nodes, enabling the Transformer encoder to encode both graph and text. GraphToken~\cite{perozzi2024let} evaluates a series of GNN models as graph encoders, including GCN, MPNN~\cite{gilmer2017neural}, GIN, Graph Transformer, HGT~\cite{hu2020heterogeneous}, etc.

    \item \textit{Text Encoding.} Due to the tremendous capability of LLMs in understanding textual information, most existing methods, such as ProteinChat~\cite{guo2023proteinchat} and DrugChat~\cite{liang2023drugchat}, directly employ LLMs as text encoders. In GraphLLM~\cite{chai2023graphllm}, the tokenizer and frozen embedding table of LLM are leveraged to obtain the representation of node text attributes, aligning with the downstream frozen LLM.

\end{itemize}
\noindent \textbf{Fusion Module.} The goal of the fusion module is to align the graph and text modalities, generating a fusion embedding as input for the LLM.
 To achieve the goal, a straightforward solution is to design a linear projection layer to directly transform the graph representation generated by GNN into an LLM-compatible soft prompt vector ~\cite{liang2023drugchat,zhang2023making,luo2023biomedgpt}.
Additionally, inspired by BLIP2's Q-Former~\cite{li2023blip}, \cite{liu2023gitmol} propose a GIT-Former, which aligns graph, image, and text with the target text modality using self-attention and cross-attention mechanisms.

In addition to the above methods, G-Retriever is proposed to integrate both explicit and implicit structural information ~\cite{he2024g}. To be specific,  GAT is employed to encode the graph structure, while representing node and relationship details through textual prompts. 
To accommodate real-world graphs with larger scales, G-Retriever introduces a RAG module specifically designed for retrieving subgraphs relevant to user queries.

\subsection{Heterophily and Generalization}
\label{sec:Heterophily and Generalization}

Despite achieving promising performance in graph tasks, GNNs exhibit several shortcomings. A notable drawback involves the inadequacy of the neighbor information aggregation mechanism, especially when dealing with heterophilic graphs. GNN performance notably diminishes when faced with instances where adjacent nodes lack similarity~\cite{song2023ordered,luan2024graph}. Additionally, GNN encounters challenges in out-of-distribution (OOD) generalization, leading to a degradation in model performance on distributions beyond the training data. This challenge is particularly prevalent in practical applications, primarily due to the inherent difficulty of encompassing all possible graph structures within limited training data. Consequently, when GNNs infer on unseen graph structures, their performance may experience a substantial decline. This reduced generalization capability renders GNNs relatively fragile when confronted with evolving graph data in real-world scenarios. For example, GNNs may encounter difficulties handling newly emergent social relationships in social networks. 

LLMs have been utilized to mitigate the above limitations. In particular, GraphText~\cite{zhao2023graphtext} effectively decouples depth and scope by encapsulating node attributes and relationships in the graph syntax tree. This approach yields superior results compared to the GNN baseline, particularly on heterophilic graphs. Chen et al.~\cite{chen2023exploring} investigate the LLM's ability to handle OOD generalization scenarios. They utilize the GOOD~\cite{gui2022good} benchmark as the criterion, and results demonstrate that LLMs exhibit promising performances in addressing OOD generalization issues. OpenGraph~\cite{xia2024opengraph} aims at solving zero-shot graph tasks across different domains. In this model, LLMs are leveraged to generate synthetic graphs of data scarcity scenarios, thereby enhancing the pre-training process of OpenGraph.

\subsection{Improving GNN Trustworthiness}
\label{sec:Improving GNN Trustworthiness}
While GNNs have achieved remarkable success in various applications like recommender systems~\cite{gao2022graph}, and drug discovery~\cite{gaudelet2021utilizing}, recent studies have found critical vulnerabilities in their trustworthiness, particularly robustness, privacy, and explainability~\cite{zhang2024trustworthy,wu2022trustworthy}. 
For example, in fraud detection systems, GNNs may fail to detect malicious activities if subtle perturbations are introduced into transaction graphs, leading to significant financial losses.
Similarly, in healthcare applications like drug discovery or disease diagnosis, the lack of explainability can hinder clinicians' trust in model predictions, potentially leading to incorrect treatment decisions. In addition, GNNs trained on sensitive social network data risk exposing private user information, violating privacy regulations, and undermining public trust.
Therefore, it is critical to ensure the trustworthiness of GNNs.

Owing to the powerful capabilities in natural language understanding, generation, and reasoning, LLMs are promising to address the trustworthiness challenges of GNNs. For instance, LLMs can enhance GNN robustness by identifying adversarial perturbations through advanced reasoning capability and improve interpretability through explainable decision-making. Below, we summarize how LLMs can be integrated with GNNs to enhance their overall trustworthiness.

\subsubsection{Robustness}
\label{sec:Robustness}
GNNs are vulnerable to adversarial attacks, where malicious perturbations to graph structures or node features can lead to incorrect predictions, raising significant concerns about their robustness~\cite{geisler2021robustness}. 
To enhance the robustness of GNNs, recent research explores leveraging LLM's advanced understanding and reasoning capabilities to improve GNNs' defenses against adversarial attacks. 
For instance, \citet{guo2024learning} employ widely used adversarial attack methods, including PGD~\cite{madry2017towards} for structural attacks and SemAttack~\cite{wang2022semattack} for textual attacks, to evaluate the robustness of integrating LLMs in graph tasks. Specifically, they explore two distinct integration strategies: LLM-as-enhancer, where the LLM generates embeddings for text attributes, and LLM-as-predictor, where the LLM directly produces predictions. Their experiments demonstrate that both approaches significantly improve robustness compared to traditional shallow models. 
To further enhance robustness, Zhang et al.~\cite{zhang2024can} propose LLM4RGNN, a framework that leverages LLMs to defend against adversarial attacks. LLM4RGNN distills GPT-4's reasoning capabilities into a local LLM through instruction tuning, enabling it to identify malicious edges in the graph. Additionally, it employs an LM-based edge predictor to detect and restore missing critical edges. By purifying the graph structure from adversarial perturbations, LLM4RGNN significantly improves the robustness of GNNs, offering a promising direction for future research.

\subsubsection{Privacy}
\label{sec:Privacy}
Privacy is a critical concern for GNNs, as they often require access to large-scale graph data containing sensitive information, such as personal relationships, health records, or financial transactions. The interconnected nature of graph data makes traditional anonymization methods insufficient, as attackers can potentially re-identify individuals or infer sensitive information through graph structure analysis~\cite{zhang2024survey}.

Recent studies demonstrate that LLMs can extract sensitive information from GNNs~\cite{xue2025trustworthy}. For instance, \citet{guan2024large} explore how LLMs can conduct link-stealing attacks by analyzing textual attributes and posterior probabilities generated by GNNs. These attacks enable LLMs to infer hidden relationships or connections between entities, even when such links are not explicitly present in the data. This highlights the significance and potential of LLMs in addressing privacy challenges in GNNs.

\subsubsection{Explainability}
\label{sec:Explainability}
A major challenge in deploying GNNs is their lack of explainability, as they often operate as ``black boxes''. This opacity limits their adoption in critical fields like healthcare, where clear reasoning behind decisions is essential. LLMs offer a promising solution to this challenge, as they excel in natural language generation and can provide step-by-step reasoning similar to human explanations. Therefore, research on enhancing GNN explainability using LLMs has gained attention~\cite{zhang2024llmexplainer}. For instance, He et al. introduce LLM-GCE~\cite{he2024explaining}, a method that leverages LLMs to generate counterfactual explanations for GNNs in molecular property prediction tasks. By creating counterfactual graph topologies from text pairs and incorporating a feedback module to reduce LLM hallucination, LLM-GCE improves the reliability and clarity of GNN explanations.

\subsection{Discussions}

The integration of LLMs into graph tasks has demonstrated great potential in enhancing the capabilities of GNNs. However, the practical deployment of such systems necessitates a comprehensive assessment of both efficiency and performance. In particular, it is essential to understand how incorporating LLMs impacts the computational efficiency of graph tasks compared to traditional GNNs, as well as the extent to which they contribute to performance improvement.

\subsubsection{Efficiency of LLM Integration in Graph Tasks}
When integrating LLMs into graph tasks, a key concern is their impact on computational efficiency. LLMs, with their large-scale architectures, can introduce significant costs in both training and inference, raising questions about their feasibility in real-time or resource-limited environments.  

Generally speaking, the efficiency impact of LLMs depends on their utilities. When LLMs are employed only to enhance feature quality without being involved in the training process, the efficiency loss is minimal. For example, TAPE~\cite{he2023harnessing} enriches node features using additional text representations generated by LLMs while keeping the LLM fixed during training and updating only the GNN parameters. Compared to methods like GLEM~\cite{zhao2022learning}, which require an expensive iterative training process between the LLM and GNN, TAPE achieves a notable speedup—e.g., a 2.88× improvement—even when using the same backbone models.  

On the other hand, approaches that integrate LLMs directly into the prediction process to overcome GNN limitations often achieve performance gains at the cost of increased inference time. As many studies do not provide detailed efficiency comparisons, it is difficult to assess the trade-offs between performance improvements and computational cost. To address this issue, GLBench~\cite{li2024glbench} has reproduced popular LLM-enhanced graph learning methods under a unified experimental setup. The results indicate that current methods exhibit significantly higher time and space complexity than traditional GNN approaches. For instance, in GLBench's experiments, GraphText~\cite{zhao2023graphtext}, which uses LLMs for prediction on the Cora dataset, had a training time exceeding \(10^3\) seconds, while a standard GCN required only about 10 seconds. This demonstrates that while LLMs can enhance performance, their integration into graph tasks comes at a substantial computational cost, highlighting the need for more efficient integration strategies to balance performance gains with practical feasibility.

\subsubsection{ Performance of LLM Integration in Graph Tasks}
The performance of LLMs on graph tasks varies depending on their role in the model. \citet{chen2023exploring} and \citet{li2024glbench} evaluated two main categories of LLM-enhanced methods: adapting LLMs to improve feature quality and directly employing them for prediction. Their findings consistently show that LLMs adapted as feature enhancers generally improve downstream task performance. For instance, 
OFA~\cite{liu2023one}
leverages LLMs to generate feature embeddings and outperforms the best GNN-based model, GCN, with improvements of 3.20\% in accuracy and 3.49\% in F1 score on the Citeseer dataset. In contrast, LLMs used as direct predictors sometimes underperform compared to traditional GNN models, suggesting that enhancing feature quality with LLMs may be a more effective and balanced strategy. Additionally, there is no clear evidence that increasing the size of LLM parameters consistently leads to better performance on graph tasks, indicating that larger models do not always yield proportional gains~\cite{chen2023exploring,li2024glbench}.

\section{Graphs for LLMs}
\label{sec:Graph-enhanced LLMs}

LLMs have demonstrated impressive language generation and understanding capabilities across various domains. Nevertheless, they still face several pressing challenges, including factuality awareness, hallucinations, limited explainability in the reasoning process, and beyond. 
To alleviate these issues, one potential approach is to take advantage of the \emph{Knowledge Graphs} (KGs), which store high-quality, human-curated factual knowledge in a structured format~\cite{chen2022knowledge}. 
Recent reviews~\cite{hu2023survey, agrawal2023can, 
pan2023unifying} have summarized the research on using KGs to enhance LMs. Hu et al.~\cite{hu2023survey} present a review on knowledge-enhanced pre-training language models for natural language understanding and natural language generation. Agrawal et al.~\cite{agrawal2023can} systematically review research on mitigating hallucination in LLMs by leveraging KGs across three dimensions: inference process, learning algorithm, and answer validation. Pan et al.~\cite{pan2023unifying} provides a comprehensive summary of the integration of KGs and LLMs from three distinct perspectives: KG-enhanced LLMs, LLM-augmented KGs, and the synergized LLMs and KGs, where LLMs and KGs mutually reinforce each other. 

In this section, we will delve into relevant research that explores the usage of KGs to achieve knowledge-enhanced language model pre-training, mitigate hallucinations, and improve inference explainability.

\subsection{KG-enhanced (L)LM Pre-training}
\label{sec:KG-enhanced (L)LM Pre-training}
While LLMs excel in text understanding and generation, they may still produce grammatically accurate yet factually incorrect information. Explicitly incorporating knowledge from KGs during LLM pre-training holds promise for augmenting LLM's learning capacity and factual awareness~\cite{sun2021ernie,fang2022knowledge,li2022oerl}. However, due to the limited research on KG-enhanced LLM pre-training, we will also cover studies on KG-enhanced pre-trained language models (PLMs), which can provide valuable insights for developing KG-enhanced LLM pre-training techniques. Existing KG-enhanced pre-training methods are generally categorized into three main approaches: modifying input data, modifying model structures, and modifying pre-training tasks.

\subsubsection{Modifying Input Data}
\label{sec:Modifying Input Data}
Several researchers investigate integrating KG knowledge by modifying input data while keeping the model architecture unchanged. For instance, Moiseev et al.~\cite{moiseev2022skill} directly train PLMs on mixed corpora consisting of factual triples from KGs and natural language texts. E-BERT~\cite{poerner2019bert} aligns entity vectors with BERT's wordpiece vector space, preserving the structure and refraining from additional pre-training tasks.  KALM~\cite{rosset2020knowledge} utilizes an entity-name dictionary to identify entities within sentences and employs an entity tokenizer to tokenize them. The input of the Transformer consists of the original word embeddings and entity embeddings. Moreover, K-BERT~\cite{liu2020kbert} integrates the original sentence with relevant triples by constructing a sentence tree, where the trunk represents the original sentence and the branches represent the triples. To convert the sentence tree into model input, K-BERT introduces both a hard-position index and a soft-position index within the embedding layer to differentiate between original tokens and triple tokens. 

\subsubsection{Modifying Model Structures}
\label{sec:Modifying Model Structures}
Some research designs knowledge-specific encoders or fusion modules to better inject knowledge into PLMs. ERNIE~\cite{zhang2019ernie} introduces a K-Encoder to inject knowledge into representations. This involves feeding token embeddings and the concatenation of token embeddings and entity embeddings into a fusion layer for generating new token embeddings and entity embeddings. In contrast, CokeBERT~\cite{su2021cokebert} extends this approach by incorporating relation information from KGs during pre-training. It introduces a semantic-driven GNN model to assign relevant scores to relations and entities based on the given text. Finally, it fuses the selected relations and entities with text using a K-Encoder similar to ERNIE. KLMO~\cite{he2021klmo} propose Knowledge Aggregator to fuse text modality and KG modality during pre-training. To incorporate the structural information in KG embeddings, KLMO utilizes KG attention, which integrates a visibility matrix with a conventional attention mechanism, facilitating interaction among adjacent entities and relations within the KG. Subsequently, the token embeddings and contextual KG embeddings are aggregated with entity-level cross-KG attention.

Several studies refrain from modifying the overall structures of the language model but introduce additional adapters to inject knowledge. To preserve the original knowledge within PLMs, Wang et al.~\cite{wang2021kadapter} propose K-Adapter as a pluggable module to leverage KG knowledge. During pre-training, the parameters of the K-Adapter are updated while the parameters of the PLMs remain frozen. KALA~\cite{kang2022kala} introduces a Knowledge-conditioned Feature Modulation layer, which functions similarly to an adapter module, by scaling and shifting the intermediate hidden representations of PLMs with retrieved knowledge representations. To further control the activation levels of adapters, DAKI~\cite{lu2021parameter} incorporates an attention-based knowledge controller module, which is an adapter module with additional linear layers.

\subsubsection{Modifying Pre-training Tasks}
\label{sec:Modifying Pre-training Tasks}
To explicitly model the interactions between text and KG knowledge, various pre-training tasks are proposed. Three major lines of work in this direction include entity-centric tasks~\cite{zhang2019ernie,yamada2020luke,xu2023kilm,shen2020exploiting,joshi2020spanbert}, relation-centric tasks ~\cite{sun2021ernie}, and beyond. 

For entity-centric tasks, ERNIE~\cite{zhang2019ernie} randomly masks some token-entity alignments and then requires the model to predict all corresponding entities based on aligned tokens. LUKE~\cite{yamada2020luke} uses Wikipedia articles as training corpora and treats hyperlinks within them as entity annotations, training the model to predict randomly masked entities. KILM~\cite{xu2023kilm} also utilizes hyperlinks in Wikipedia articles as entities. However, it inserts entity descriptions after corresponding entities, tasking the model with reconstructing the masked description tokens rather than directly masking entities. In addition to predicting masked entities, GLM~\cite{shen2020exploiting} further introduces a distractor-suppressed ranking task. This task leverages negative entity samples from KGs as distractors, enhancing the model's ability to distinguish various entities.

Relation-centric tasks are also commonly utilized in KG-enhanced PLMs. For instance, JAKET~\cite{yu2022jaket} proposes relation prediction and entity category prediction tasks for enhancing knowledge modeling. Dragon~\cite{yasunaga2022deep} is pre-trained in a KG link prediction task. Given a text-KG pair, the model needs to predict the masked relations in KG and the masked tokens in the sentence. ERICA~\cite{qin2020erica} introduces a relation discrimination task aiming at semantically distinguishing the proximity between two relations. Specifically, it adopts a contrastive learning manner, wherein the relation representations of entity pairs belonging to the same relations are encouraged to be closer.

In the context of LLMs, Jiang et al.~\cite{jiang2024efficient} propose a triples-to-text generation task to enhance the models' understanding of KG information and employ the LoRa technique to achieve domain knowledge pre-learning.

\subsection{KG-enhanced LLM Inference}
\label{sec:KG-enhanced LLM Inference}

Knowledge within KGs can be dynamically updated with relatively low computational cost, whereas updating the knowledge in LLMs typically requires adjusting model parameters, which is both resource-intensive and time-consuming. Therefore, many studies opt to integrate KGs during the inference stage of LLMs. This approach allows LLMs to leverage up-to-date information without the need for costly re-training. Additionally, the ``black-box'' nature of LLMs makes it challenging to understand how a model arrives at a specific prediction or generates a particular response. Given the structured and fact-based representation of knowledge in KGs, incorporating them during inference can enhance the transparency and explainability of LLM outputs, thereby reducing the risk of hallucinations and improving the overall reliability of the model. Generally speaking, existing KG-enhanced inference methods can be broadly categorized into two types: \textit{KG-enhanced LLM Inference through In-Context Learning} and \textit{KG-enhanced LLM Inference through Fine-tuning}. 

\subsubsection{Enhancing by In-Context Learning}
\label{sec:icl}
In-Context Learning provides a flexible and efficient way to integrate external knowledge into LLMs without modifying their internal parameters. By incorporating KG knowledge into prompts, LLMs can dynamically access structured, up-to-date information during inference, enhancing their ability to generate accurate and reliable outputs. Researchers have explored various methods to effectively inject KG information into LLMs through in-context learning. For instance, KAPING~\cite{baek2023knowledge} directly inputs relevant triplets in the format ``(head entity, relation, tail entity)'' into the LLM. However, \citet{wu2023retrieverewriteanswer} argue that such a representation neglects the gap between KG structures and natural language, potentially hindering the LLM's understanding. To address this problem, they employ ChatGPT to convert KG triplets into natural language (e.g., transforming (China, capital, Beijing) into ``China's capital is Beijing'') and fine-tune open-source LLMs with these natural language expressions, enabling them to better interpret KG knowledge. In addition, several studies also explored representing KG information as reasoning paths, allowing LLMs to process KG knowledge step by step~\cite{luo2023reasoning,feng2023knowledge,wang2023keqing}. For example, Knowledge Solver~\cite{feng2023knowledge} first provides the LLM with a head entity, a query, and candidate tail entities. Subsequently, the LLM selects an appropriate tail entity from the candidates, which serves as the new head entity for subsequent iterations. This process repeats until the entity answering the query is identified.

\subsubsection{Enhancing by Fine-tuning}
\label{sec:finetuning}

While in-context learning provides a flexible way to integrate KG knowledge into LLMs, the limited context window of LLMs restricts their ability to process large amounts of KG information efficiently, potentially leading to suboptimal performance. 
To address this issue, researchers have explored fine-tuning the LLM to enhance its ability to interpret and utilize KG information. 
A widely adopted method is employing instruction tuning to adapt LLMs for better utilization of KG knowledge.
For instance, ~\citet{tian2024kg} propose inserting KG-Adapter layers within the LLM architecture and jointly performing instruction tuning to effectively integrate KG information. These KG-Adapter layers facilitate the integration of LLM and KG semantics through bi-directional cross-attention mechanisms, enabling the model to better capture the relationships between textual and structured knowledge.
In contrast to enhancing the LLM's understanding of the KG, KG-Agent~\cite{jiang2024kg} fine-tunes the LLM to actively retrieve and reason information directly from the KG. Specifically, it employs programmatic language to structure the multi-hop reasoning process over the KG and creates a code-based instruction dataset to fine-tune the LLM. This approach extends the LLM's functionality beyond knowledge integration, enabling it to execute complex reasoning tasks over structured knowledge.

Given the substantial computational costs of fine-tuning LLMs, recent studies introduce the graph encoder to encode KG knowledge. Instead of updating the entire LLM, these methods generate KG embeddings that serve as soft prompts, guiding the LLM to leverage KG knowledge without extensive fine-tuning.
For example, GNP~\cite{tian2023graph} employs a GNN model to encode retrieved KG information, which is then processed through a cross-modality pooling module and a domain projector to align the KG embeddings with the LLM's embedding space. These aligned embeddings are subsequently combined with input text embeddings for inference. Similarly, GraphLLM~\cite{chai2023graphllm} utilizes a graph transformer~\cite{dwivedi2020generalization} to encode graph information, which is prefixed to the input prompt and fed into the LLaMA model. 
In contrast to these static integration approaches, InfuserKI~\cite{wang2024infuserki} introduces a novel method to dynamically decide whether to incorporate KG embeddings into the LLM. This two-step process begins with knowledge detection, where InfuserKI identifies unknown knowledge by querying the LLM with questions derived from KG triplets. When the LLM lacks relevant information, InfuserKI adaptively injects the necessary knowledge into the LLM. This design allows InfuserKI to balance the integration of new knowledge with the preservation of the LLM's existing capabilities, providing a more flexible and efficient solution for KG-LLM integration.

\section{Applications}
\label{sec:application}
In this section, we will present practical applications that demonstrate the potential and value of Graph ML and LLMs. As shown in Table~\ref{tab:llm-enhanced graph}, recommender systems, knowledge graphs, AI for science, and robot task planning emerge as the most prevalent domains. We will provide a comprehensive summary of each of these applications.

\subsection{Recommender Systems}
\label{sec:Recommender Systems}
Recommender systems leverage user historical behaviors to predict items that users are likely to appreciate~\cite{fan2018deep,fan2019deep_daso,fan2019deep_dscf,liu2024score}. Graphs play a crucial role in recommender systems, wherein users and items can be regarded as nodes and collaborative behaviors such as clicks and purchases can be viewed as edges. 
Recently, an increasing amount of research is exploring the use of LLMs for direct recommendation~\cite{yin2023heterogeneous,tang2023one,dai2023uncovering,wang2024rethinking,qu2024tokenrec,qu2024ssd4rec,ning2024cheatagent} or leveraging LLMs to enhance graph models or datasets for recommendation tasks~\cite{weillmrec,xi2023towards,ren2023representation,wu2023leveraging,hao2023lkpnr}.

For directly using LLMs as recommendation models, liu et al.~\cite{liu2023chatgpt} construct task-specific prompts to evaluate ChatGPT's performance on five common recommendation tasks, encompassing rating prediction, sequential recommendation, direct recommendation, explanation generation, and review summarization. Bao et al.~\cite{bao2023tallrec} employ prompt templates to guide LLM to decide whether the user will like the target item based on their historical interactions and perform instruction tuning on the LLM to improve its recommendation capability.

For using LLMs to enhance traditional recommendation methods or datasets, KAR~\cite{xi2023towards} leverages LLMs to generate factual knowledge of items and reasoning basis of user preferences; these knowledge texts are then encoded into vectors and integrated into existing recommendation models. Methods like LLM-Rec~\cite{lyu2023llmrec}, RLMRec~\cite{ren2023representation}, and LLMRec~\cite{weillmrec} enrich recommendation datasets by incorporating LLM-generated descriptions. In contrast, Wu et al.~\cite{wu2023leveraging} utilize LLMs to condense recommendation datasets, in which  LLMs are employed to synthesize a condensed dataset for the content-based recommendation, aiming at addressing the challenge of resource-intensive training on large datasets.

While the previously discussed methods have explored utilizing LLMs for certain recommendation tasks or domains, an emerging research direction aims to develop foundation models for recommendation. Tang et al.~\cite {tang2023one} propose an LLM-based domain-agnostic framework for sequential recommendation. Their approach integrates user behavior across domains, and leverages LLMs to model user behaviors based on multi-domain historical interactions and item titles. Hua et al.~\cite{hua2023up5} attempt to address the potential unfairness of recommender systems introduced by LLM bias. They propose a Counterfactually Fair Prompting method to develop an unbiased foundation model for recommendation. To summarize the progress in the area of recommendation foundation model, Huang et al.~\cite{huang2024foundation} provide a systematic overview of the existing approaches, categorizing them into three main types: language foundation models, personalized agent foundation models, and multi-modal foundation models.

\subsection{Knowledge Graphs}
\label{sec:Knowledge Graphs}

LLMs with robust text generation and language understanding capabilities have found extensive applications in KG-related tasks, including KG completion~\cite{yang2023cpkgc,zhang2023making,yao2023exploring}, KG question answering~\cite{kim2023kggpt,luo2023chatkbqa,wu2023retrieverewriteanswer,feng2023knowledge,jiang2023structgpt}, KG reasoning~\cite{luo2023chatrule} and beyond. 
Meyer et al.~\cite{meyer2023developing} introduce LLM-KG-Bench, a framework that automatically evaluates the model's proficiency in KG engineering tasks such as fixing errors in Turtle files, facts extraction, and dataset generation. 
KG-LLM~\cite{yao2023exploring} is proposed to evaluate LLMs' performance on KG completion, including triple classification, relation prediction, and link prediction tasks. 
Kim et al.~\cite{kim2023kggpt} propose KG-GPT, using LLMs for complex reasoning tasks on knowledge graphs. ChatKBQA~\cite{luo2023chatkbqa} introduces a generate-then-retrieve framework for LLMs on knowledge base question answering.
Wu et al.~\cite{wu2023retrieverewriteanswer} present a KG-enhanced LLM framework for KG question answering, which involves fine-tuning an LLM to convert structured triples into free-form text, enhancing LLMs' understanding of KG data. 
The successful application of LLMs in tasks such as KG construction, completion, and question answering offers robust support for advancing the understanding and exploration of KGs. 

Drawing inspiration from foundation models in language and vision, researchers are delving into the development of foundation models tailored for KGs. These models aim to generalize to any unseen relations and entities within KGs. Galkin et al.~\cite{galkin2023towards} propose Ultra, which learns universal graph representations by leveraging interactions between relations. This study is based on the insight that those interactions remain similar and transferable across different datasets.

\subsection{AI for Science}
\label{sec:AI for Science}
The rapid advancement of AI has led to an increasing number of studies leveraging AI to assist scientific research~\cite{tang2023general,cao2022identifying,fan2025computational}. 
Recent research has applied LLMs and Graph ML methods
for scientific purposes, such as drug discovery, molecular property prediction, and material design.
Notably, these applications encompass scenarios involving graph-structured data. 

The molecular graph is a way of representing molecules, where the nodes represent atoms, and the edges represent the bonds between the atoms.
With the emergence of LLMs, researchers have explored their performance in tasks related to molecular graphs. Methods like MolReGPT~\cite{li2023empowering} and GPT-MolBERTa~\cite{balaji2023gptmolberta} adopt a similar approach, converting molecular graphs into textual descriptions using SMILES language.
They create prompts based on SMILES data, asking the LLM to provide detailed information about functional groups, shapes, chemical properties, etc. This information is then used to train a smaller LM for molecular property prediction. 
In contrast to methods directly using LLMs for prediction, ReLM~\cite{shi2023relm} first uses GNNs to predict high-probability candidate products, and then leverages LLMs to make the final selection from these candidates.
In addition to the above research, LLMs are further utilized in drug discovery and materials design. Bran et al.~\cite{bran2023chemcrow} present ChemCrow, a chemistry agent integrating LLMs and 18 specialized tools for diverse tasks across drug discovery, materials design, and organic synthesis.
InstructMol~\cite{cao2023instructmol} presents a two-stage framework for aligning language and molecule graph modalities in drug discovery. Initially, the framework maintains the LLM and graph encoder parameters fixed, focusing on training the projector to align molecule-graph representations. Subsequently, instruction tuning is conducted on the LLM to address drug discovery tasks. Zhao et al.~\cite{zhao2024chemdfm} propose ChemDFM, the first dialogue foundation model for chemistry. Trained on extensive chemistry literature and general data, ChemDFM exhibits proficiency in various chemistry tasks such as molecular recognition, molecular design, and beyond.

\subsection{Robot Task Planning}
\label{sec:Robot Task Planning}
Robot task planning aims to decompose tasks into a series of high-level operations for the step-by-step completion by a robot~\cite{galindo2008robot}. During task execution, the robot needs to perceive information about the surrounding environment, typically represented using \emph{scene graphs}. In a scene graph, nodes represent scene objects like people and tables, while edges describe the spatial or functional relationships between objects.
Enabling LLMs for robot task planning crucially depends on how to represent the environmental information in the scene graph.

Many studies have explored using textual descriptions of scene information and constructing prompts for LLMs to generate task plans.  Chalvatzaki et al.~\cite{chalvatzaki2023learning} introduce the Graph2NL mapping table, representing attributes with different numerical ranges using corresponding textual expressions. For instance, a distance value greater than 5 is represented as ``distant'', and smaller than 3 is represented as ``reachable''. 
SayPlan~\cite{rana2023sayplan} describes the scene graph in JSON as a text sequence, iteratively invoking LLM to generate plans and allowing for self-correction. Zhen et al.~\cite{zhen2023robot} propose an effective prompt template, Think\_Net\_Prompt, to enhance LLM performance in task planning. In contrast to methods that rely on language to describe scene graph information, GRID~\cite{ni2023grid} employs the graph transformer to encode scene graphs. 
It utilizes cross-modal attention to align the graph modality with user instruction, ultimately outputting action tokens through a decoder layer. 
The powerful understanding and reasoning 
capabilities of LLMs showcase significant potential in robot task planning. However, as task complexity increases, the search space explosively expands, posing a challenge in efficiently generating viable task plans with LLMs. 

\section{Future Directions}
\label{sec:future_work}

In this survey, we have thoroughly reviewed the latest developments of Graph ML in the era of LLMs, revealing significant advancements and potential in this field. 
As this research direction is still in the exploratory stage, future directions in this field can be diverse and innovative. 
Therefore, in this section, we delve into several potential future directions of this promising field.

\subsection{Retrieval-Augmented Generation with Graphs}
\label{sec:GraphRAG}
Despite the notable achievements of LLMs, they have been proven to suffer from hallucinations and lack of up-to-date knowledge~\cite{luo2023reasoning}.
Recently, Retrieval-Augmented Generation (RAG) has emerged as a promising technique to overcome these limitations~\cite{zhu2024realm,borgeaud2022improving,asai2023self,fan2024survey}.
Typical studies such as RETRO~\cite{borgeaud2022improving} and REALM~\cite{zhu2024realm} retrieve relevant fragments from external knowledge databases to guide LLMs in generating outputs.
However, these vanilla RAG methods often neglect structural information in knowledge, which limits their effectiveness in downstream tasks.
To solve these problems, recent studies focus on incorporating structured factual knowledge retrieved from graphs~\cite{he2024g,luo2023reasoning,mavromatis2024gnn,wang2025knowledge,han2024retrieval}.
Unlike traditional RAG methods, these graph RAG approaches provide rich relational and structural information that is crucial for improving LLM performance.
For example, G-retriever~\cite{he2024g} retrieves subgraphs related to the query to assist LLMs, effectively mitigating issues like hallucination and lack of knowledge.
Nevertheless, bridging the gap between unstructured text generation and structured graph reasoning poses a challenge, while the efficiency of retrieving knowledge from large-scale graphs is another pressing concern.
Solving these challenges presents unique opportunities for advancing the capabilities of LLMs, especially on graph tasks.

\subsection{Generalization and Transferability}
\label{sec:Generalization and Transferability}
While Graph ML has deployed for various graph tasks, a notable problem is their limited capacity for generalization and transferability across different graph domains~\cite{zhang2023large}.
Different from fields such as NLP and CV, where data often adhere to a uniform format (e.g., a sequence of tokens or a grid of pixels), graphs can be highly heterogeneous in nature.
This heterogeneity manifests in varying graph sizes, densities, and types of nodes and edges, which presents a significant challenge in developing a universal model capable of performing optimally across various graph structure data. 
Currently, LLMs have demonstrated great potential in improving the generalization ability of the graph model. For example, OFA~\cite{liu2023one} provides a solution for classification tasks across several certain domains.
Nevertheless, there is still scarce exploration of the generalizability of Graph ML methods compared to LLMs.
Therefore, future research should aim to develop more adaptable and flexible models that can effectively apply learned patterns from one graph type, such as social networks, to another, like molecular structures, without extensive retraining.

\subsection{Trustworthiness}
\label{sec:future-Trustworthiness}
The recent applications of LLMs for Graph ML have significantly enhanced graph modeling capabilities and broadened their utility in various fields.
Despite these advancements, with the growing reliance on these models, it is important to ensure their trustworthiness, particularly in critical areas like healthcare, finance, and social network analysis~\cite{liu2021trustworthy,fan2022comprehensive}.
Robustness is fundamental in safeguarding the models against adversarial attacks, ensuring consistent reliability.
Explainability is essential for users to understand and trust the decisions made by these models. 
Fairness is crucial for the model's ethical and effective use in various applications. 
Privacy is important for legal compliance and key to maintaining user trust.
Therefore, the development of trustworthy LLMs on graphs must be equipped with \emph{Robustness}\&\emph{Safety}, \emph{Explainability}, \emph{Fairness}, and \emph{Privacy}, ensuring their safe and effective use in various applications.

\subsubsection{Robustness\&Safety}
\label{sec:Robustness-Safety}
Recently, integrating LLMs into Graph ML has shown promising performance in various downstream tasks, but they are also highly vulnerable to adversarial perturbations, raising significant concerns about their robustness and safety. 
To enhance the resilience of these models, some studies add adversarial perturbations to GNNs~\cite{fan2023jointly,fan2021attacking} or LLMs~\cite{jain2023baseline,bespalov2023towards} for adversarial training. 
However, these methods may not be effective for the new paradigm of Graph ML integrating LLMs, as vulnerabilities can arise from both graphs, such as graph poisoning attacks~\cite{chen2022understanding,zou2021tdgia} and graph modification attacks~\cite{dai2018adversarial,zugner2018adversarial}, and from the language model, like prompt attacks~\cite{wei2023jailbroken} and misleading text data~\cite{zhang2023certified}. To address these issues, more sophisticated detection and defense mechanisms need to be developed by considering both the intricacies of LLMs and graphs to ensure the comprehensive safety and robustness of Graph ML.

\subsubsection{Explainability}
\label{sec:future-Explainability}
Nowadays, LLMs are increasingly employed in Graph ML across various applications, such as recommender systems~\cite{geng2022recommendation,devlin2018bert} and molecular discovery~\cite{li2023empowering,su2022molecular}. However, due to privacy and security concerns, an application provider may prefer to provide an API version without revealing the architecture and parameters of the LLM, such as with ChatGPT. This lack of transparency can make it challenging for users to understand the model's results, leading to confusion and dissatisfaction.
Therefore, it's important to enhance the explainability of Graph ML, especially with LLMs. 
Existing studies have focused on improving interpretability in GNNs~\cite{ying2019gnnexplainer,luo2020parameterized,zheng2023towards}. For example, GNNExplainer~\cite{ying2019gnnexplainer} generates explanations by identifying subgraphs and node features critical to predictions. Furthermore, PGExplainer~\cite{luo2020parameterized} improves the generalization and performance of GNNExplainer by introducing trainable parameters.
Despite these advances, these methods are tailored to traditional GNN architecture and may struggle to generate human-intuitive explanations for complex tasks.
Notably, owing to the powerful reasoning and interpretive capabilities of LLMs, they are promising to provide better explainability in graph-related tasks.
For example, P5~\cite{geng2022recommendation} can provide reasons for its recommendations in recommendation tasks. Future efforts should be directed toward making the inner workings of these models more transparent and explainable to better comprehend their decision-making processes.

\subsubsection{Fairness}
\label{sec:Fairness}
As LLMs become prevalent in enhancing Graph ML, concerns about their fairness are growing. Fairness is crucial to ensure these models operate without biases or discrimination, especially when dealing with complex, interconnected graph data~\cite{fan2022comprehensive}. Recent studies demonstrate that both language models~\cite{liu2020does,zhang2022fairness} and GNN models~\cite{chen2023fairly} can potentially be discriminatory and unfair~\cite{li2023survey}.
Therefore, it is necessary to maintain fairness in both textual and graph contexts.
To enhance the fairness of LLMs, recent studies include retraining strategies that adjust model parameters for unbiased outputs\cite{webster2020measuring}, implementing alignment constraints~\cite{guo2022auto}, and adopting contrastive learning to diminish bias in model training~\cite{he2022mabel}. 
Concurrently, studies like FairNeg~\cite{chen2023fairly} also explore improving the fairness of recommendation data. 
Despite these efforts, achieving fairness in Graph ML methods is still a significant challenge that needs further exploration.

\subsubsection{Privacy}
\label{sec:future-Privacy}
Privacy is a critical issue in Graph ML, particularly given the risk of these models inadvertently leaking sensitive information contained in graph data~\cite{carlini2021extracting,huang2022large,zhang2024graph}.
For example, Graph ML integrated with LLMs could potentially expose private user data, like browsing histories or social connections when generating outputs.
This concern is especially pressing in highly data-sensitive areas such as healthcare or finance. To mitigate these privacy risks, ~\cite{li2023privacy} introduces Privacy-Preserving Prompt Tuning (RAPT) to protect user privacy through local differential privacy.
Future exploration in LLM-enhanced Graph ML should also focus on integrating privacy-preserving technologies like differential privacy and federated learning to strengthen data security and user privacy.

\subsection{Efficiency}
\label{sec:future-Efficiency}
While LLMs have proven effective in building more powerful Graph ML methods, their operational efficiency, particularly in processing large and complex graphs, is still a significant challenge~\cite{xue2023efficient,qu2024survey}. 
For example, the use of APIs like GPT4 for large-scale graph tasks can lead to high costs under current billing models. 
Additionally, deploying open-source LLMs (e.g., LLaMa) for parameter updates or just inference in local environments demands substantial computational resources and storage. 
Therefore, enhancing the efficiency of LLMs for graph tasks remains a critical issue. Recent studies introduce techniques like LoRA~\cite{hu2021lora} and QLoRA~\cite{dettmers2023qlora} to fine-tune LLM parameters more efficiently.
Furthermore, model pruning~\cite{ma2023llm,xia2023sheared} is also a promising method to increase efficiency by removing redundant parameters or structures from LLMs, thereby simplifying their application in Graph ML.

\subsection{Multi-modal Graph Learning}
\label{sec:Multi-modal Graph Learning}
Recent LLMs have shown significant potential in advancing Graph ML. 
Many efforts have been made to transform graph data into formats suitable for LLM input, such as tokens or text~\cite{zhao2023graphgpt,chen2023exploring,wang2023can}.
However, many nodes in graphs are enriched with diverse modalities of information, including text, images, and videos.
Understanding this multi-modal data can potentially benefit graph learning. 
For example, on social media platforms, a user's post could include textual content, images, and videos, all of which are valuable for comprehensive user modeling.
Given the importance of multi-modal data, a promising direction for future research is to empower LLMs to process and integrate graph structure with multi-modal data.
Currently, TOUCHUP-G~\cite{zhu2023touchup} makes an initial exploration of multi-modal ( i.e., texts, images) in graph learning.
In the future, we expect the development of a unified model capable of modeling universal modalities for more advanced Graph ML methods.

\section{Conclusion}
\label{Conclusion}
In this survey, we have thoroughly reviewed the recent progress of Graph ML in the era of LLMs, an emerging field in graph learning.
We first review the evolution of Graph ML, and then delve into various methods of LLMs enhancing Graph ML.
Due to the remarkable capabilities in various fields, LLMs have great potential to enhance Graph ML.
We further explore the augmenting of LLMs with graphs, highlighting their ability to enhance LLM pre-training and inference. 
Additionally, we demonstrate their potential in diverse applications such as molecule discovery, knowledge graphs, and recommender systems.
Despite their success, this field is still evolving and presents numerous opportunities for further advancements.
Therefore, we further discuss several challenges and potential future directions.
Overall, our survey aims to provide a systematic and comprehensive review to researchers and practitioners, inspiring future explorations in this promising field.

\section{Review Methodology}
To provide a systematic and reproducible survey for graph machine learning in the era of large language models, we adopt a structured literature review methodology. Specifically, we mainly focus on studies from 2018 to 2025 to capture recent advancements in this rapidly evolving field. Studies are mainly selected from top conferences in data mining (KDD, WWW, WSDM, CIKM, ICDE, etc.), NLP (ACL, EMNLP, NAACL, etc.), and AI/ML conferences (AAAI, IJCAI, NeurIPS, ICLR, ICML, etc.) as well as top journals including TKDE, TOIS, TKDD, and TIST.
However, as this field is emerging and rapidly evolving, several studies in this area are still under review and in preprint. Therefore, we also cover the studies that are in arXiv and OpenReview to include the latest work not yet formally published.

\bibliographystyle{ACM-Reference-Format}
\bibliography{sample-acmsmall}

\end{document}